\documentclass[journal]{IEEEtran}
\usepackage{color}
\usepackage{bbding}
\usepackage{pifont}
\usepackage{amssymb}
\usepackage{amsthm}
\usepackage{booktabs}
\usepackage{multirow} 
\usepackage{setspace}
\usepackage{epsfig}
\usepackage{graphicx}
\usepackage{subfigure}
\usepackage{algorithm}
\usepackage{algpseudocode}
\usepackage{url}
\usepackage{amsmath}
\usepackage{bm}
\usepackage{tablefootnote}
\usepackage{float}

\newcommand{\etal}{\textit{et~al.}}
\newcommand{\ie}{\textit{i.e.}}
\newcommand{\eg}{\textit{e.g.}}

\usepackage{array}
\newcolumntype{L}[1]{>{\raggedright\let\newline\\\arraybackslash\hspace{0pt}}m{#1}}
\newcolumntype{C}[1]{>{\centering\let\newline\\\arraybackslash\hspace{0pt}}m{#1}}
\newcolumntype{R}[1]{>{\raggedleft\let\newline\\\arraybackslash\hspace{0pt}}m{#1}}

\usepackage{xcolor}

\ifCLASSINFOpdf
\else
\fi
\begin{document}
	\title{LithoHoD: A Litho Simulator-Powered Framework for IC Layout Hotspot Detection}
	
	\author{Hao-Chiang~Shao,~\IEEEmembership{Member,~IEEE}, Guan-Yu~Chen, Yu-Hsien~Lin, Chia-Wen~Lin,~\IEEEmembership{Fellow,~IEEE}, Shao-Yun~Fang,~\IEEEmembership{Member,~IEEE}, 
	Pin-Yian~Tsai, and~Yan-Hsiu~Liu% <-this % stops a space
	\thanks{Manuscript received on 10 April 2024; revised 2 August 2024; accepted 15 September 2024.  Date of publication Month Date, 2024; date of current version Month Date, 2024. This work was supported in part by the National Council of Science and Technology, Taiwan, under Grants NSTC 112-2634-F-002-005 and NSTC 112-2221-E-005-080, and in part by United Microelectronics Corporation. The associate editor coordinating the review of this manuscript and approving it for publication was Dr. Laleh Behjat. (Corresponding Author: Chia-Wen Lin)}
	\thanks{Hao-Chiang Shao is with the Institute of Data Science
            and Information Computing, National Chung Hsing University, Taichung 402202, Taiwan. (e-mail:shao.haochiang@gmail.com)}
	\thanks{Guan-Yu Chen and Yu-Hsien~Lin are with the Department of Electrical Engineering, National Tsing Hua University, Hsinchu 300044, Taiwan.}		
	\thanks{Chia-Wen Lin is with the Department of Electrical Engineering and the Institute of Communications Engineering, National Tsing Hua University, Hsinchu 300044, Taiwan. (e-mail: cwlin@ee.nthu.edu.tw)}
	\thanks{Shao-Yun Fang is with the Department of Electrical Engineering, National Taiwan University of Science and Technology, Taipei 106335, Taiwan. (e-mail: syfang@mail.ntust.edu.tw)}
	\thanks{Pin-Yian Tsai is with the Product Engineering Department, United Microelectronics Corporation, Hsinchu 300094, Taiwan. (e-mail: pin\_yian\_tsai@umc.com)}
        \thanks{Yan-Hsiu Liu is with the Development of Smart Manufacturing, United Microelectronics Corporation, Hsinchu 300094, Taiwan. (e-mail: cecil\_liu@umc.com)}
	\thanks{Color versions of one or more of the figures in this paper are available online at http://ieeexplore.ieee.org.}
	}
	

	\markboth{IEEE Transactions on Computer-Aided Design of Integrated Circuits and Systems,~Vol.~x, No.~x, Month~2024}%
	{Shell \MakeLowercase{\textit{et al.}}: Bare Demo of IEEEtran.cls for IEEE Journals}



	\maketitle
	
	\begin{abstract}
        Recent advances in VLSI fabrication technology have led to die shrinkage and increased layout density, creating an urgent demand for advanced hotspot detection techniques. 
However, by taking an object detection network as the backbone, recent learning-based hotspot detectors learn to recognize only the problematic layout patterns in the training data. This fact makes these hotspot detectors difficult to generalize to real-world scenarios. We propose a novel lithography simulator-powered hotspot detection framework to overcome this difficulty. 
Our framework integrates a lithography simulator with an object detection backbone, merging the extracted latent features from both the simulator and the object detector via well-designed cross-attention blocks. 
Consequently, the proposed framework can be used to detect potential hotspot regions based on i) the variation of possible circuit shape deformation estimated by the lithography simulator, and ii) the problematic layout patterns already known. To this end, we utilize RetinaNet with a feature pyramid network as the object detection backbone and leverage LithoNet as the lithography simulator.
Extensive experiments demonstrate that our proposed simulator-guided hotspot detection framework outperforms previous state-of-the-art methods on real-world data. 

	\end{abstract}
	

	% \fi
	\IEEEpeerreviewmaketitle

%------------------------------------------------------------------------	
	\section{Introduction}
	\label{sec01:intro}
	With the rapid advances in the semiconductor industry, process nodes are transitioning from the tens to the single digits, resulting in increased transistor density and making identifying defect patterns on wafers more challenging. This poses an issue for designers when they attempt to correct their layout design and mitigate defects. 
To this end, hotspot detection in integrated circuit (IC) fabrication is a crucial process that entails identifying and analyzing specific areas within a chip design susceptible to potential manufacturing or operational challenges. These ``hotspots'' denote regions where particular conditions or patterns might give rise to defects, compromised performance, or other undesirable outcomes throughout the fabrication process or in the operation of the final IC.

The significance of hotspot detection lies in its pivotal role during the design and verification phases of IC development, ensuring the creation of high-quality and dependable semiconductor devices. This intricate task employs specialized tools and algorithms to scrutinize the layouts and designs of ICs, searching for patterns or configurations that could lead to issues such as timing violations, excessive power consumption, or challenges associated with fabrication processes like lithography.
Early identification of hotspots in the design phase empowers engineers and manufacturers to implement corrective measures promptly. These measures may involve optimizing the design, adjusting manufacturing processes, or introducing enhancements to alleviate potential problems. This proactive approach enhances the overall yield, reliability, and performance of ICs, ultimately creating more efficient and resilient electronic devices.

Traditionally, designers employ a scanning electron microscope (SEM) to scrutinize post-process images of a wafer to identify defects. However, with denser designs comes the need for more high-resolution SEM images to meet defect detection requirements, leading to substantial time and labor costs. Although pattern-matching approaches can quickly detect hotspot patterns on a layout design, they are still not robust against novel design patterns.

Recently, several learning-based methods have been proposed to address this problem by adopting convolutional neural network (CNN) based object detection models \cite{jiang2021efficient,geng2020hotspot,yang2019layout,chen2019faster,gai2021flexible,zhu2021hotspot, asp-dac2023}. 
By regarding a layout hotspot region as an object, the hotspot detection problem can be recast as an object detection issue, where the solver aims to identify different objects, \ie, hotspots, in an input image along with their classification probabilities and bounding-box regions. This consideration has led to the recent design trend in this type of approach. 
For example, Chen \etal~\cite{chen2019faster} devised a hotspot detector based on Faster-RCNN~\cite{ren2015faster}. Their model employs an inception-based feature extractor and a clip proposal network to generate region proposals, improving recall performance, reducing false alarms, and increasing computation speed.

\begin{figure*}[!t]
     \centering
    \includegraphics[width=0.80\textwidth]{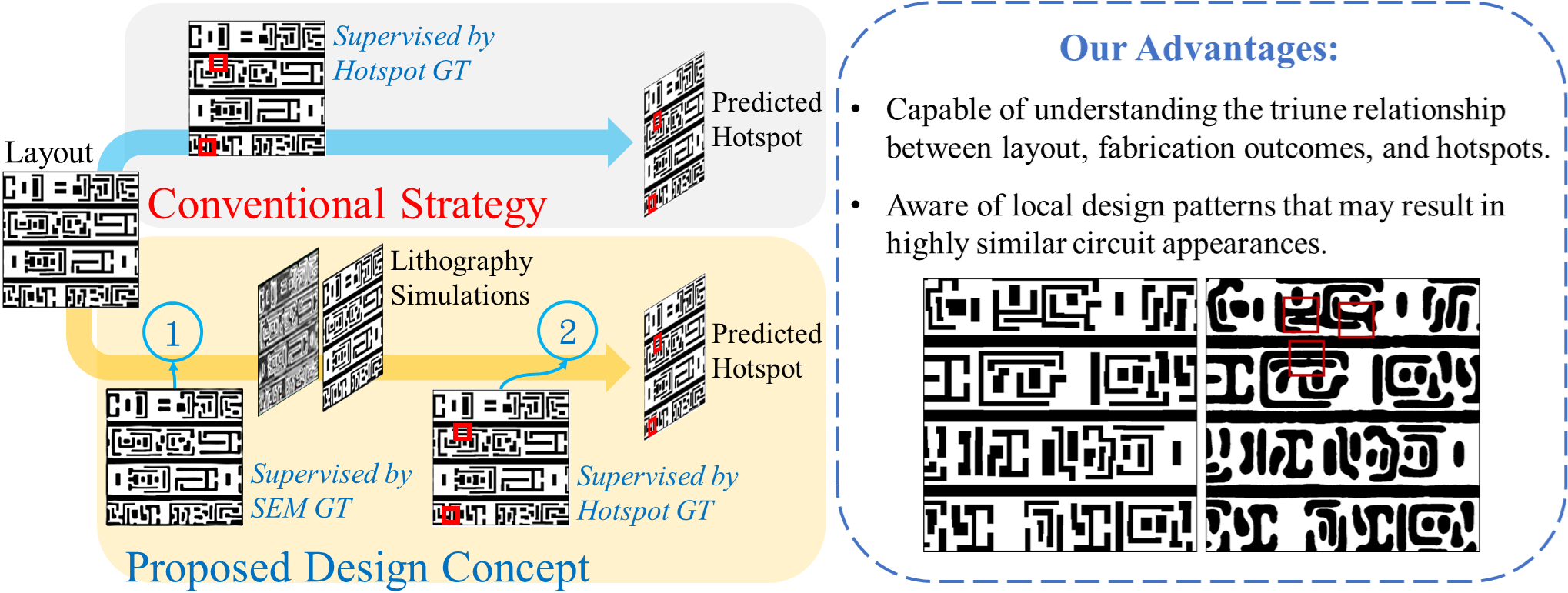} 
     \caption{
    Differences in the design concepts between traditional hotspot detectors and ours. While a traditional hotspot detector is trained as a naive hotspot pattern recognizer with the supervision of hotspot ground-truths, our design aims to learn the triune relationship among layout, fabrication outcomes, and hotspots. This design enables \textbf{LithoHoD} to recognize local design patterns that may lead to highly similar circuit appearances of hotspots. As a result, \textbf{LithoHoD} functions as a versatile hotspot detector capable of identifying not only layout defect patterns but also considering circuit shape deformations due to a fabrication process to pinpoint potential hotspots. Hotspot regions in this figure are indicated with red boxes. }
     \label{fig:partition}
\end{figure*}

\iffalse
However, by taking a general object detection network as the backbone, recent learning-based hotspot detectors can only learn to recognize problematic layout patterns similar to those already appearing in the training data. This fact makes such hotspot detectors difficult to generalize to real-world layout patterns unseen during training. 
To address this issue, we propose a \textbf{Litho} (lithography) simulator-powered \textbf{Ho}tspot \textbf{D}etector (\textbf{LithoHoD}) for real-world layout hotspot detection tasks. 
Fig.~\ref{fig:partition} illustrates the design goal \textcolor{blue}{and the design concept} of the proposed framework. Specifically, we deem that a hotspot detector should not only recognize layout defect patterns but also consider the circuit shape distortions caused by the fabrication process for identifying potential hotspot regions. To this end, our framework integrates a learning-based lithography simulator with an object detection network by fusing their feature tensors via primarily a well-designed cross-attention block consisting of several attention modules. As a result, our method can detect potential hotspot regions based on i) the possible variations of circuit shape distortion estimated by the litho simulator, and ii) the problematic layout patterns already learned. 
\fi

%\textcolor{blue}{
However, by using a general object detector as the backbone, conventional learning-based hotspot detectors only learn to recognize problematic layout patterns similar to the hotspot ground-truths in the training data. 
This makes conventional hotspot detectors struggle to generalize to real-world applications, where different layouts may result in similar hotspot circuit appearances due to fabrication process variations. 
To address this issue, we propose a \textbf{Litho} (lithography) simulator-powered \textbf{Ho}tspot \textbf{D}etector (\textbf{LithoHoD}) for real-world layout hotspot detection tasks. 

Fig.~\ref{fig:partition} illustrates the differences between the design concepts of LithoHoD and conventional hotspot detectors. 
Traditional methods rely solely on object detectors to identify regional structures of layout patterns prone to causing hotspots. In contrast, LithoHoD integrates a lithography simulator, using its predicted deformation map to improve feature embedding for the object detection network, thus enhancing hotspot identification.
%
%Specifically, we design our hotspot detector to recognize both layout defect patterns and potential hotspot regions by considering the circuit shape distortions caused by the fabrication process. To this end, our framework integrates a learning-based lithography simulator with an object detection network by fusing their feature tensors via well-designed cross-model feature fusion modules. 
%
Specifically, LithoHoD learns the triune relationship among fabrication imperfections (\textit{i.e.}, the predicted deformation maps), layout patterns, and hotspot ground truths during the training process.
LithoHoD integrates a learning-based lithography simulator with an object detection network by fusing their feature tensors via well-designed cross-model feature fusion modules. This design enables LithoHoD to recognize different local design patterns that may lead to similar defect appearances after fabrication, thereby improving detection accuracy.
As a result, by learning the triune relationship between i) circuit layouts, ii) circuit fabrication outcomes, and iii) hotspot ground truths, our method can detect potential hotspot regions based on the possible variations in circuit shape distortions estimated by the lithography simulator and the problematic layout patterns already learned.

The main contribution of this paper is twofold. First, we propose a simulator-powered framework for layout hotspot detection. With the aid of a pretrained lithography simulator, the proposed framework is able to detect potential hotspot regions instead of identifying a hotspot pattern already seen. Second, we propose a cross-attention block that guides the fusion of the feature tensor extracted by the lithography simulator with that extracted by common object detection networks. Our extensive experiments demonstrate the effectiveness of the proposed framework.

\iffalse\begin{figure}[!t]
     \centering
     \includegraphics[width=0.45\textwidth]{figure/fig01_hotspot_intro.jpg} 
     \caption{Illustration of the design goals of the proposed \textbf{LithoHoD}. Left: a layout clip. Right: a binarized SEM image of the product IC corresponding to the given layout.  \textbf{LithoHoD} not only recognizes layout defect patterns but also considers circuit shape distortion to identify potential hotspots. In this figure, hotspot regions are labeled manually within the red boxes.}
     \label{fig:partition}
\end{figure}
\fi

%------------------------------------------------------------------------	
	\section{Related Work}
	\label{sec02:review}
	
In this work, we propose formulating IC layout hotspot detection as a supervised learning-based visual object detection problem based on the layout shape features, involving the interactions between the neighboring layout designs and the lithography-induced shape deformation. This section reviews the most relevant methods in learning-based hotspot detection, object detection, and lithography simulation.

\subsection{Learning-based Hotspot Detection Methods}
\label{section:machine_learning_based_hotspot_detection}

Since Yu \etal~\cite{yu2013machine} proposed a pioneering machine learning-based hotspot detection method in 2013, learning-based approaches have already demonstrated the capability in layout hotspot detection tasks. For example, Matsunawa \etal~\cite{matsunawa2015new} presented a hotspot detection framework utilizing an Adaboost classifier and a simplified feature extraction scheme. Later, Yang \etal~develop a CNN-based hotspot detection method based on features derived by a traditional static image compression scheme~\cite{yang2019layout}. 
In Yang \etal's approach, a layout is first transformed into frequency domain features patch-wisely by using a discrete cosine transform followed by a zig-zag scan to form a feature tensor; then, the resulting feature tensor is fed into a LeNet-styled neural network to derive the detection result. Additionally, Chen \etal~\cite{chen2019faster} proposed a hotspot detection method based on Faster-RCNN, employing an inception module as its feature extraction backbone. Despite achieving good detection accuracy, the above three methods incur high computational time costs due to the design of the network framework. 
Hence, Gai \etal~\cite{gai2021flexible} proposed a computationally efficient lightweight network model to reduce computation. However, this method is specialized as a binary hotspot detector and is thus unable to recognize different hotspot categories. As a result, this method needs to be re-trained to adapt to hotspot patterns of different types.

Additionally, the data security issue in layout hotspot detection problems has been addressed recently. Specifically, Lin \etal~\cite{lin2022lithography} employed heterogeneous federated learning, facilitating the adaptation of the global model to local data heterogeneity while preserving the privacy of local data and preventing the leakage of layout designs. This is particularly crucial for advanced process nodes in today's competitive landscape as companies strive to safeguard their designs from external exposure.

\subsection{Object detection networks}
\label{section:Retinanet}

Object detection has been among the most widely utilized computer vision techniques since anchor-based deep object detection networks (\textit{e.g.}, R-CNN~\cite{girshick2014rich}) were introduced. Deep object detection networks are generally categorized into two types, namely one-stage and two-stage detectors, based on whether a pre-generated region proposal is employed in the object detection process. Typically, two-stage detectors achieve higher accuracy, but often lead to longer execution times due to repeated computations. The representative two-stage detectors belong to the R-CNN family~\cite{girshick2014rich,girshick2015fast,ren2015faster}. Using a fixed number of predefined anchors as candidate regions for analysis, R-CNN~\cite{girshick2014rich} extracts deep features from these candidates and classifies them via a support vector machine. Fast R-CNN~\cite{girshick2015fast} further improves computation efficiency by converting an input image into a base feature tensor at the outset. It then crops regional features from the feature tensor based on predefined candidate regions for subsequent classification and bounding box regression tasks. Subsequently, Faster R-CNN~\cite{ren2015faster} then introduces a region proposal network (RPN) for suggesting region proposals at runtime, improving recognition performance and saving computation costs by performing computations on only those regions identified by the RPN.

In contrast, one-stage detectors perform region localization and object classification concurrently, making them faster but less accurate than two-stage detectors. Representative one-stage object detectors include the YOLO family~\cite{yolov1,yolov2,yolov3, yolov4}, SSD~\cite{liu2016ssd}, and RetinaNet~\cite{lin2017focal}. Particularly, RetinaNet addresses the data imbalance problem, caused by the uneven distribution between foreground and background objects, using focal loss. It also utilizes the Feature Pyramid Network (FPN)~\cite{lin2017feature} to recognize objects across different sizes and observation scales. These improvements enable RetinaNet to run as fast as one-stage detectors while maintaining accuracy comparable to two-stage detectors. Therefore, building upon RetinaNet, we devise our layout hotspot detection model.

\subsection{Lithography Simulators}

\begin{figure*}[t]
    \centering
    \includegraphics[width=0.95\textwidth]{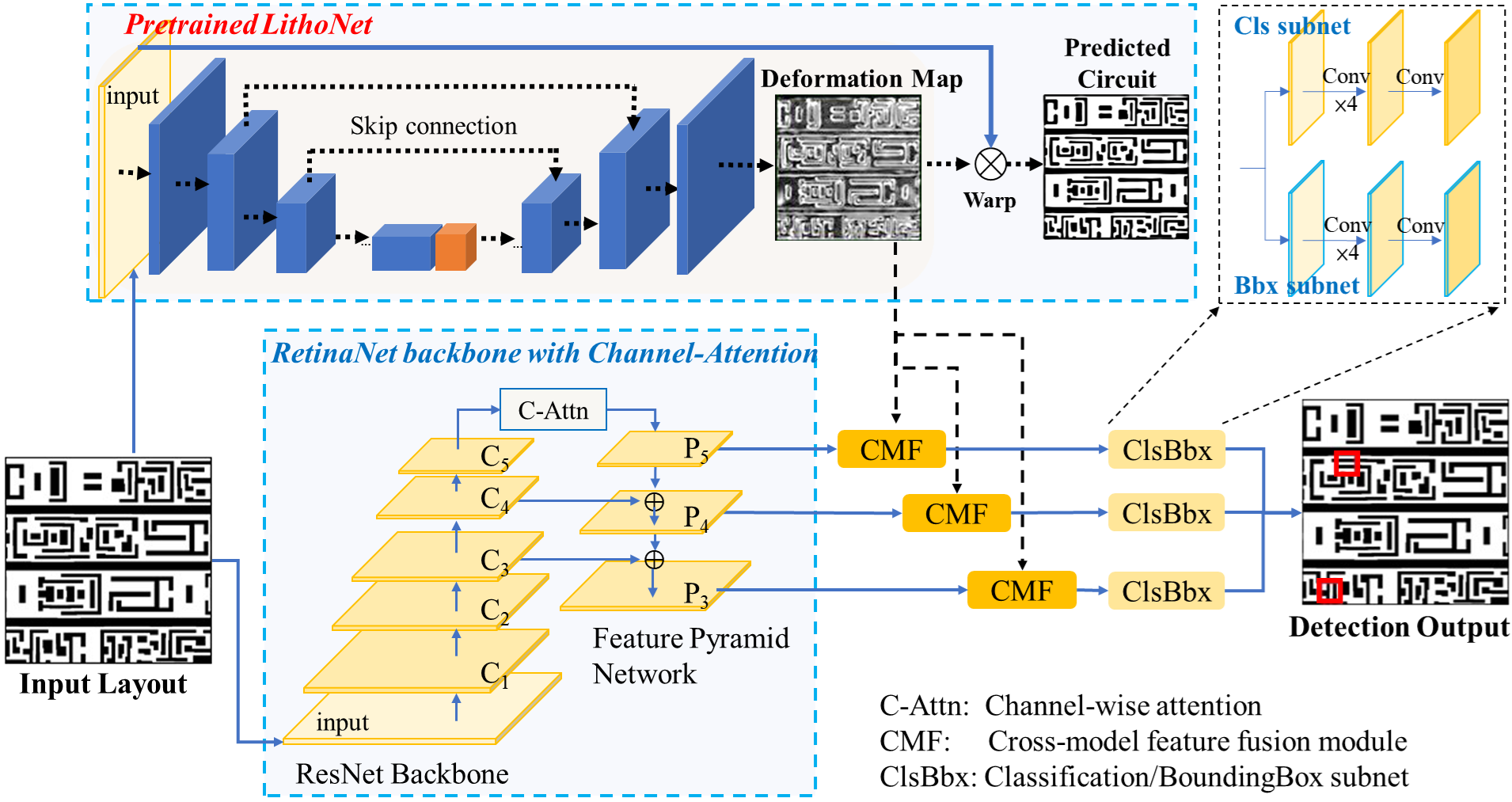}%{figure/fig02_archi_F.png}
    \caption{Framework of the proposed LithoHoD. This framework comprises two primary functional components: a pretrained lithography simulator (LithoNet) and a trainable object detector (Channel-attention-enhanced RetinaNet). This design includes a channel-wise attention (C-Attn) module and three cross-model feature fusion (CMF) modules. The C-Attn module enhances the representability of the feature tensor used in the RetinaNet backbone. In contrast, the CMF modules facilitate the integration of the shape-deformation feature  $f_\mathrm{de}$ with detection features $f_{\mathrm{py},l}$ in three distinct observation scales extracted by the feature pyramid network (FPN). Here, the tensor $\mathrm{C}_l$ corresponds to the output of the $l$-th convolution stage, and its dimension is $2^l$ times lower than the original input. $\mathrm{P}_l$ is the tensor derived by the pyramid structure with the same spatial dimension as $C_l$. Note that, i) the FPN in RetinaNet omits the $P_2$ and $P_1$ layers in its design, ii) our implementation removes the $P_6$ and $P_7$ layers of the original FPN used in RetinaNet, and iii) the architecture of the detector module \textbf{ClsBbx}, consisting of a classification subnet and a box regression subnet, is detailed in Table \ref{table:detectormodule}.}
    \label{fig:proposed_arch}
\end{figure*}

Traditionally, two approaches have been employed to obtain a post-process aerial image of an IC layout. One involves a real fabrication procedure to acquire a SEM (scanning electron microscope) image of a product IC. In contrast, the other utilizes lithography simulators, such as ICWB (IC WorkBench) by Synopsys. However, both of these conventional methods are expensive and time-consuming. Consequently, several neural network-based lithography simulation models were proposed to address this challenge~\cite{watanabe2017accurate,GANOPC2018yang,ye2019lithogan,asp-dac2023,lithonet}. 
In particular, LithoNet~\cite{lithonet}, a data-driven learning-based lithography simulator, can predict the circuit shape deformation due to the lithography and etching processes involved in IC fabrication with varying fabrication parameters. LithoNet learns the circuit shape correspondence from paired layout and SEM images to estimate a deformation map through a multi-scale generator, as illustrated in the upper branch of Fig.~\ref{fig:proposed_arch}. This deformation map and the latent features derived by LithoNet capture crucial information such as circuit shapes and deformation variances. Hence, LithoNet is further employed to construct a novelty-based active-learning framework for best updating a pre-trained lithography simulator~\cite{shao2023keeping}.
Building upon the capabilities of LithoNet and leveraging its deformation map as a guidance feature, we propose a model-guided lithography hotspot detection method in this paper, namely \textbf{LithoHoD}.

%------------------------------------------------------------------------	
	\section{Lithography Simulator-Powered Hotspot Detector}
	\label{sec03:method}

\begin{figure*}[t]
    \centering
    \includegraphics[width=0.96\textwidth]{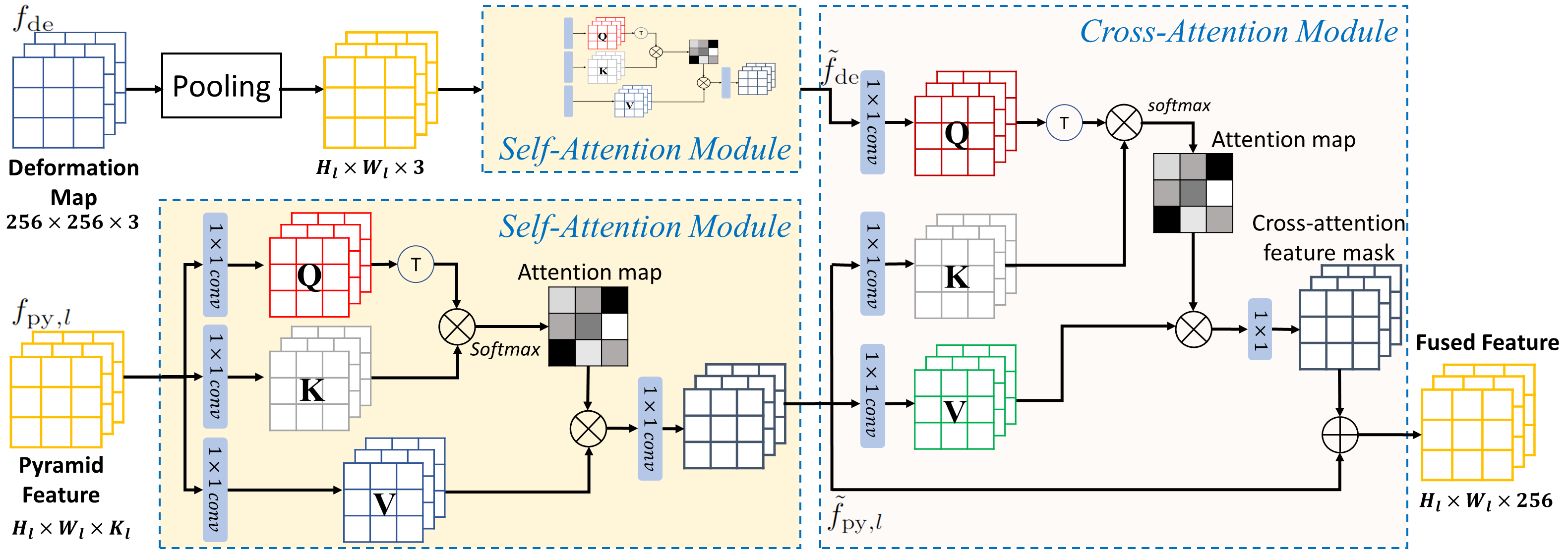} 
    \caption{Proposed Cross-Model Feature Fusion (CMF) module. This module is designed to generate a fused model-guided feature for layout hotspot detection. To this end, CMF comprises two self-attention modules and one cross-attention module. Specifically, one self-attention module enhances the shape-deformation feature (\textit{i.e.}, the deformation map $f_{\mathrm{de}}$) generated by the pre-trained lithography simulator.  Besides, the other self-attention module processes the pattern feature (\textit{i.e.,} the pyramid feature $f_{\mathrm{py},l}$ for object detection) derived from the feature pyramid network of the channel-attention-enhanced RetinaNet backbone.  The two enhanced and adapted feature tensors are then fed into the cross-attention module for cross-model feature fusion. Note that the two self-attention modules share the same architecture, and 
    the dimension of each feature tensor used in the CMF module is shown in Table \ref{tab:tensordimension}.
    }
    \label{fig:attentionmodules}
\end{figure*}

\subsection{Overview}

We devise a lithography simulator-assisted framework for layout hotspot detection. 
The proposed framework, as depicted in Fig.~\ref{fig:proposed_arch}, comprises four components: i) a pretrained lithography simulator, ii) an object detection network, iii) a channel-wise attention (C-Attn) module within the object detector backbone, and iv) our devised cross-model feature fusion (CMF) modules, each comprising two self-attention modules and one cross-attention module. 
C-Attn aims to enhance the coarsest features extracted from the layout pattern by the object detector. Additionally, the CMF module is designed to fuse tensors of different domains and dimensions: the deformation features generated by the lithography simulator and the pattern features generated by the object detector backbone. The self-attention mechanism in the CMF module enhances the representability of its input tensor by exploring the dependency between two different spatial positions within the tensor, whereas the cross-attention mechanism transfers the simulator's prediction of circuit shape deformation to the object detection network, enabling the recognition of potential hotspot areas. This design enables the proposed framework to detect hotspots based on the learnable relationship between layout patterns, lithography distortions, and defect patterns in the training dataset. 
 
LithoHoD requires a more comprehensive training dataset than previous hotspot detection methods. This dataset includes i) layout patterns, ii) hotspot ground truths, and iii) SEM images of product ICs, where the third one is not used in the traditional methods. The training process for LithoHoD consists of two stages: pre-training and main training. 
In the pre-training stage, the lithography simulator (LithoNet) is trained using pairs of layout patterns and their corresponding SEM images. 
 During the main training stage, the pretrained LithoNet, serving as the upper branch of LithoHoD, predicts circuit shape deformation maps (\textit{i.e.}, local shape deformation features) caused by an imperfect fabrication process. These deformation features are then fused with the layout shape features extracted by the object detector (RetinaNet) in the lower branch, which is trained on pairs of layout patterns and hotspot ground truths. This fusion, achieved through the self- and cross-attention mechanisms of the Cross-Model-Feature (CMF) module, enables more accurate hotspot detection by leveraging the complementary features provided by the lithography simulator. 

%\textcolor{blue}{
%Finally, note that in LithoHoD's framework, the lithography simulator guides the hotspot detection process through our proposed cross-model feature fusing (CMF) module, via which  the deformation map derived by LithoNet is fused with the object detection feature tensors derived by RetinaNet. The labeled hotspot ground-truths are never altered during the training period. }
In the following subsections, we will elaborate on each component of our framework design.

\subsection{LithoNet: Assessing lithography distortion}

Hotspot detection aims to identify and analyze chip design areas prone to potential manufacturing or operational issues, \eg, distortions caused by the lithographic process.
To identify potential hotspots, including those caused by lithography gaps and those sensitive to the lithographic process~\cite{yu2012accurate}, we utilize a pretrained data-driven lithography simulator LithoNet that we proposed in~\cite{lithonet} in the proposed hotspot detection framework to assess lithography distortions. This design allows our detector to consider factors such as lithography distortions and fabrication parameters, which have not yet been explored in existing CNN-based hotspot detection approaches. 
Specifically, LithoNet can generate a lithography prediction $\mathcal{J'}$ and a shape deformation map $\mathcal{M}$ for an input layout clip by learning the shape correspondence between the training layout patterns $\mathcal{S}$ and their lithography ground-truth $\mathcal{J}$ (\textit{i.e.}, the corresponding SEM image). %This deformation map records the displacement vectors of individual pixels, offering features for representing pixel-wise lithography distortions. These features can then be used to guide the layout hotspot detection process. 
Thus, the predicted deformation map $\mathcal{M}: \mathbf{R}^2 \rightarrow \mathbf{R}^2$ records the pixel-wise displacement vectors between individual pixels $(x_\mathcal{S}, y_\mathcal{S})$ on a layout image $\mathcal{S}$ and the pixel position $(x_\mathcal{J}, y_\mathcal{J})$ on the associated SEM image $\mathcal{J}$. Consequently, the warping operator, illustrated in the upper half of Fig. \ref{fig:proposed_arch}, transforms an input layout shape into its corresponding fabricated circuit shape based on the predicted displacement vectors from the predicted deformation map $\mathcal{M}$. 
As a result, the deformation map, which records the displacement vectors of individual pixels, provides features that represent pixel-wise lithography distortions. This deformation map $\mathcal{M}$ is then considered as a deformation feature $f_\mathrm{de}$ that can potentially guide the layout hotspot detection process.

However, the deformation map predicted by LithoNet only consists of pixel-wise displacement vectors, which is insufficient to capture the interactions between the layout contours within their local neighborhood. Therefore, we enhance the deformation map by using a self-attention module to assess the inter-pixel dependencies. As reported in~\cite{shao2023keeping}, the self-attention module enhances a feature tensor by assessing how the feature of the $i$-th element of the tensor contributes to that of the $j$-th element. This enhancement process facilitates the learning of local interactions between shapes within a neighborhood and results in the deformation feature $f_\mathrm{de}$, representing the attention the pretrained lithography simulator provides. The deformation feature $f_\mathrm{de}$ can then be utilized to assist the layout hotspot detection process through a cross-attention module.

Note that the pre-trained LithoNet learns to predict lithography distortions from training data pairs consisting of i) a layout pattern and ii) the product IC's SEM images of the given layout, using the loss terms described in \cite{lithonet}. LithoNet is essentially a CNN aiming to predict the fabricated circuit shape of an input layout. In other words, the pre-trained LithoNet is knowledgeable about the circuit shape distortion caused by a fabrication process. Additionally, the labeled hotspot ground truths required for LithoHoD's main training stage will not be altered by the pretrained LithoNet.

\subsection{Object Detection Backbone}

The backbone of the proposed hotspot detection framework is a modified RetinaNet, which is enhanced with a channel-wise attention module.
RetinaNet~\cite{lin2017focal} is a widely adopted object detection system known for its high detection accuracy and  low computational cost. 
The feature extraction backbone of RetinaNet is built upon ResNet~\cite{he2016deep} and FPN (feature pyramid network) \cite{lin2017feature}. In the original RetinaNet design, the ResNet backbone converts an input image into multi-stage feature tensors. Subsequently, FPN utilizes these multi-stage feature tensors as inputs to generate feature tensors of different observation scales for the subsequent object recognition and classification stage. This design enables RetinaNet to learn multi-resolution features that are robust to the changes in the target object size.

In our design, we make several modifications to RetinaNet. First, we added the channel-wise attention (CA) module, used in the CBAM (Convolutional Block Attention Module)~\cite{woo2018cbam}, to the connection between the deepest feature tensor outputted by ResNet and the coarsest-level input of FPN. This channel-wise attention module allows our network to learn the importance of individual channels in the high-level feature tensor, which is particularly helpful for characterizing the shape properties of layout patterns composed exclusively of polygons, as opposed to real-world objects, for subsequent hotspot detection~\cite{woo2018cbam}.
Second, to efficiently merge feature tensors of layout patterns, we modified the network architecture of RetinaNet \cite{lin2017focal} and FPN \cite{lin2017feature} by removing the $P_6$ and $P_7$ layers from their original designs\footnote{Please refer to the third-party implementation of RetinaNet:  https://github.com/yhenon/pytorch-retinanet/blob/master/retinanet/model.py}. Instead, we employed a shallower architecture comprising only $P_3$, $P_4$, and $P_5$ layers\footnote{The FPN of RetinaNet does not include the $P_2$ and $P_1$ layers in its design.}. In other words, we construct a feature pyramid with levels $P_3$ through $P_5$, where $P_l$ has a resolution of $2^l$ times lower than the original input image, where $l$ represents the pyramid level. This modification not only reduces the network parameters but also prevents the use of feature tensors that correspond to a representation that is too coarse to capture the local interaction of layout contours. Hence, this modification can avoid possible overfitting caused by binary and polygonal layout patterns. 
Third, we incorporate an additional self-attention module (within the cross-model feature fusion module, which will be introduced later) to enhance the multi-resolution feature tensors generated by FPN. This self-attention module improves the multi-resolution features by focusing on local interactions between layout contours within a neighborhood.

\subsection{Cross-Model Feature Fusion (CMF) Module}

Inspired by the self-attention module proposed in~\cite{zhang2019self} and its extension in the application to layout novelty detection \cite{shao2023keeping}, we devise a Cross-Model Feature Fusion (CMF) module. As shown in Fig.~\ref{fig:attentionmodules}, this modules consists of two self-attention blocks and a meticulously designed cross-attention block, which enables the integration of the lithography simulator with the object detector. This integration facilitates a lithography simulator-guided hotspot detection process.

One main issue in merging the deformation map 
$f_\mathrm{de}$ extracted by the lithography simulator and the $l$-th level pyramid feature $f_{\mathrm{py,}l}$ (\textit{i.e.}, the pattern feature shown in Fig.~\ref{fig:attentionmodules}) obtained from the modified RetinaNet for object detection lies in aligning these two feature tensors. To this end, we first reduce the spatial dimension of deformation map $f_\mathrm{de}$ 
using global average pooling\footnote{
Deformation map $\mathcal{M}$ records the pixel-wise displacement vectors of a $W \times H$ input layout image. Downsampling (average-pooling) the deformation map to fit the different spatial dimensions of the feature tensors ($P_3$--$P_5$) from different layers of the FPN does not lead to information loss. Specifically, the deformation map is resized to $h_i\times w_i$ via average pooling to guide the hotspot detector layer with a $h_i\times w_i$ feature tensor $P_i$ by feature fusion.
As a result, the hotspot detection process is conducted in a multi-resolution fashion, where the resized deformation features are fused with feature tensors $P_3$--$P_5$ of the coarse-to-fine detection branches in parallel at different observation scales. This process thus does not lead to information loss.} to match that of $f_{\mathrm{py,}l}$. Then, the spatially aligned deformation map and the pyramid feature individually pass through two independent self-attention modules to align their channel dimensions, allowing them to be effectively fused using the cross-attention block we devise. 
Fig.~\ref{fig:attentionmodules} illustrates the block diagrams of our cross-model feature fusion (CMF) module that consists of two self-attention blocks and one cross-attention block in our implementation. Considering that each layer of the pyramid features is individually fed into an object detection subnetwork of RetinaNet, three CMF modules aim to independently integrate the deformation features into the $P_3$, $P_4$, and $P_5$ layers of the pyramid features, as illustrated in Fig.~\ref{fig:proposed_arch}. Notice that we use $\tilde{f}_\mathrm{de}$ and $\tilde{f}_{\mathrm{py,}l}$ to denote the tensors enhanced and aligned by the self-attention modules, respectively.

\textbf{Cross-Attention Block.} We adopt the architecture of the self-attention module from SA-LithoNet~\cite{shao2023keeping} and repurpose it into our cross-attention block. 
While the self-attention block enhances the representability of its input feature tensor by providing it with an attention map on a wider neighborhood based on the input itself, the cross-attention block improves its target tensor based on an attention map recording the contribution the $i$-th element of the guiding tensor brings to the $j$-th element of the target tensor. As a result, the guiding tensor in our cross-attention block is enhanced with a wide-range representative feature for capturing the fabrication-induced shape deformation of a circuit. The proposed cross-attention block takes the enhanced deformation feature $\tilde{f}_\mathrm{de}$ as the input of the \textit{query} ($\mathbf{Q}$) branch and the enhanced pyramid detection feature $\tilde{f}_{\mathrm{py,}l}$ as the inputs of the \textit{key} ($\mathbf{K}$) and \textit{value} ($\mathbf{V}$) branches. 
The cross-attention computation can be formulated in the following matrix form: 
\begin{equation}
    %\left[
    \begin{array}{ccc}
         \mathbf{Q}=\mathbf{W}^T_q \, \tilde{\mathbf{f}}_\mathrm{de}, & \mathbf{K}=\mathbf{W}^T_k \, \tilde{\mathbf{f}}_{\mathrm{py,}l}, & \mathbf{V}=\mathbf{W}^T_v \, \tilde{\mathbf{f}}_{\mathrm{py,}l}, \\
    \end{array} 
    %\right] 
\label{eq:selfAtt_qkv}
\end{equation}
where $\tilde{\mathbf{f}}_\mathrm{de}$ and $\tilde{\mathbf{f}}_{\mathrm{py,}l}$ represent the deformation and pyramid feature tensors enhanced by the self-attention modules, respectively, and $\mathbf{W}^T_q$, $\mathbf{W}^T_k$, and $\mathbf{W}^T_v$ denote $1 \times 1$ convolution kernels. 
Via tensor reshaping, the dimensions of $\tilde{\mathbf{f}}_\mathrm{de}$ and $\tilde{\mathbf{f}}_{\mathrm{py,}l}$ become 
$C_1\times (H \cdot W)$ and $C_2\times (H \cdot W)$, respectively, where $H$ and $W$ are the height and the width of the tensor. Meanwhile, the dimensions of $\mathbf{W}^T_q$, $\mathbf{W}^T_k$, and $\mathbf{W}^T_v$ are ${C_1\times C}$, $C_2 \times C$, and $C_2 \times C$, respectively. Notably, in our implementation, $C=32$, $C_1=3$, and $C_2=256$.

As a result, the attention map of the cross-attention block in the Fig.~\ref{fig:attentionmodules} can be derived by 
\begin{equation}
    \beta_{j,i} = \frac{e^{s_{ij}}}{\sum_{i=1}^{HW}\sum_{j=1}^{HW} e^{s_{ij}}}\mbox{,}
    \label{eq:XA_beta}
\end{equation}
where $s_{ij} = \mathbf{q}_{i}^{T}\mathbf{k}_{j}$; $\mathbf{q}_{i}$ and $\mathbf{k}_{j}$ are %$C\times 1\times 1$
$C\times 1$
sub-tensors, and $\beta_{j,i}$ represents the normalized attention (\ie, the dependency) in the $j$-th location contributed by the $i$-th region. 

Consequently, the output cross-attention feature mask $\mathbf{M}~=~(\mathbf{m}_{1}, \mathbf{m}_{2}, \dots, \mathbf{m}_j, \dots,\mathbf{m}_{HW})$ is a $C_2 \times (W \cdot H)$ tensor with  $\mathbf{m}_{j}$ defined as  
\begin{equation}
    \mathbf{m}_{j} = \mathbf{W}^T_m \, \sum_{i=1}^{HW} \beta_{j,i} \mathbf{v}_{i} \mbox{,}
    \label{eq:attentionoutput}
\end{equation}
where $\mathbf{v}_{i}$ denotes the $i$-th %$C_2\times 1 \times 1$ 
$C_2\times 1 $ 
sub-tensor of the $C_2\times(W\cdot H)$ \textit{value} map, and $\mathbf{W}^T_m$ denotes a $1 \times 1$ convolution kernel with a dimension of $C\times C_2$.

Consequently, the final feature tensor $\mathbf{f}_\mathrm{XA}(\mathbf{x})$ enhanced by this cross-attention module becomes 
\begin{equation}
    \mathbf{f}_\mathrm{XA} = \xi \, \mathbf{M} + \tilde{\mathbf{f}}_{\mathrm{py,}l} \mbox{,}
    \label{eq:featureXA}
\end{equation}
where $\xi$ is a learnable parameter, initialized as $1$. Note that the dimension of $\mathbf{f}_\mathrm{XA}$ is identical to the cross-attention feature mask $\mathbf{M}$, \textit{i.e.,} $C_2 \times (W \cdot H)$, in the matrix representation, and $\mathbf{f}_\mathrm{XA}$ is a $H\times W\times C_2$ tensor in our implementation.

This cross-attention block can learn the spatial dependency between the two input feature tensors and thus extract a tensor more representative than its input for hotspot detection. We take this cross-attention detection feature $\mathbf{f}_\mathrm{XA}$ in (\ref{eq:featureXA}) as the input for the subsequent detector module depicted in Subsection \ref{subsec:detector}.

\textbf{Self-Attention Block.} Our self-attention Block is structurally a simplified version of the cross-attention block described above. Concisely speaking, taking the self-attention block for the deformation map $\mathbf{f}_\mathrm{de}$ provided by the lithography simulator for example, its \textit{query}, \textit{key} and \textit{value} maps are defined as $\mathbf{Q}_\mathrm{self}=\mathbf{W}^T_\mathrm{Qs} \, \mathbf{f}_\mathrm{de}$, $\mathbf{K}_\mathrm{self}=\mathbf{W}^T_\mathrm{Ks} \, \mathbf{f}_\mathrm{de}$, and $\mathbf{V}_\mathrm{self}=\mathbf{W}^T_\mathrm{Vs} \, \mathbf{f}_\mathrm{de}$, respectively. The attention map of the self-attention module is defined in the same way as in (\ref{eq:XA_beta}). As a result, the self-attention block outputs a tensor, whose each $C_s \times 1 \times 1$ sub-sensor is defined similarly to (\ref{eq:attentionoutput}).  Table~\ref{tab:tensordimension} summarizes the dimensions of individual feature tensors used in the proposed cross-attention block.

\iffalse
\begin{table}[t]
    \centering
    \caption{Dimension of each feature tensor used in the proposed cross-domain feature fusion (CMF) module}
    \footnotesize    
    \begin{tabular}{|c|cc|c|}
    \hline
         Tensors & \multicolumn{2}{|c|}{Self-Attention Module} & Cross-Attention Module\\
          & $f_\mathrm{de}$ & $f_{\mathrm{py,}l}$ & \\
         \hline
         Input& H$\times$W$\times$3 & H$\times$W$\times$256 & $f_\mathrm{de}$: H$\times$W$\times$3\\
              & & & $f_{\mathrm{py,}l}$: H$\times$W$\times$256\\
        \cline{2-4}
         \textbf{Q} & H$\times$W$\times$3 & H$\times$W$\times$16 & H$\times$W$\times$32\\
         \textbf{K} & H$\times$W$\times$3 & H$\times$W$\times$16 & H$\times$W$\times$32\\
         \textbf{V} & H$\times$W$\times$3 & H$\times$W$\times$16 & H$\times$W$\times$32\\
         Attention map & HW$\times$HW & HW$\times$HW & HW$\times$HW\\
         \cline{2-4}
         Output& H$\times$W$\times$3 & H$\times$W$\times$256 & H$\times$W$\times$256\\
         \hline
    \end{tabular}
    \label{tab:tensordimension}
\end{table}
\fi
\begin{table}[t]
    \centering
    \caption{Dimension of each feature tensor used in the proposed cross-model feature fusion (CMF) module}
    \scriptsize    
    \begin{tabular}{|c||cc|c|}
    \hline
         \textbf{Tensors} & \multicolumn{2}{c|}{\textbf{Self-Attention Blocks}} & \textbf{Cross-Attention Block}\\
          & $f_\mathrm{de}$ & $f_{\mathrm{py,}l}$ & \\
         \hline
         \hline
         Input& 256$\times$256$\times$3 & H$_l\times$W$_l\times$K$_l$ & $f_\mathrm{de}$: H$_l\times$W$_l\times$3\\
         After pooling     & H$_l\times$W$_l\times$3 & -\,- & $f_{\mathrm{py,}l}$: H$_l\times$W$_l\times$256\\
        \hline %\cline{2-4}
         \textbf{Q} & H$_l\times$W$_l\times$3 & H$_l\times$W$_l\times$16 & H$_l\times$W$_l\times$32\\
         \textbf{K} & H$_l\times$W$_l\times$3 & H$_l\times$W$_l\times$16 & H$_l\times$W$_l\times$32\\
         \textbf{V} & H$_l\times$W$_l\times$3 & H$_l\times$W$_l\times$16 & H$_l\times$W$_l\times$32\\
         Attention map & H$_l$W$_l\times$H$_l$W$_l$ & H$_l$W$_l\times$H$_l$W$_l$ & H$_l$W$_l\times$H$_l$W$_l$\\
         \hline %\cline{2-4}
         Output& H$_l\times$W$_l\times$3 & H$_l\times$W$_l\times$256 & H$_l\times$W$_l\times$256\\
         \hline
         \multicolumn{4}{l}{$^*$W$_5$=H$_5$=16, K$_5$=2048; W$_4$=H$_4$=32, K$_4$=1024; and, W$_3$=H$_3$=64, K$_3$=512.}
    \end{tabular}
    \label{tab:tensordimension}
\end{table}
\begin{table}[t]
\caption{Architectures of Bounding-Box Regression Submodule and Classifier Submodule}
\centering
\scriptsize
\begin{tabular}{|c|| c | c | c |}
\hline
 & \multicolumn{3}{c|}{Functional Layers} \\ \cline{2-4}
Submodule & Conv2D+ReLU & Final Conv. & Final Activation \\  \hline \hline
BBox-Regression  & $\times4$ & $\times1$ & --  \\ \hline
Classification   & $\times4$ & $\times1$ & Sigmoid  \\ \hline
\end{tabular}
\label{table:detectormodule}
\end{table}

\subsection{Detector Module}
\label{subsec:detector}

The detector module is responsible for inferring the object category and locating the object bounding box from its input feature tensor at the end of the object detection network. A detector module usually consists of two sub-networks: a classification subnet and a box regression subnet. This subsection explains the detector configuration of the proposed LithoHoD. 
The architectures of our classification and box regression subnets are shown in Table \ref{table:detectormodule}.

\textbf{Classification Subnet.} The classification subnet aims to predict the category probability of a target object, along with a hotspot area. Given that our proposed simulator-guided hotspot detector is adopted from a RetinaNet~\cite{lin2017focal} backbone, we highlight the key settings below.

First, because the proposed LithoHoD adopts a five-level feature pyramid network instead of the original seven-level architecture, it only utilizes three different sizes of anchors:  $32\times32$ for $P_3$, $64\times64$ for $P_4$, and $128\times128$ for $P_5$, respectively.
Second, the hyper-parameter \textbf{SCALES} is reconfigured to be $[0.25, 0.5, 1.0, 2.0]$. This hyper-parameter controls the usage of either enlarged or shrunk anchors, enabling the hotspot detector to identify hotspot regions of variable sizes. 
Third, the hyper-parameter \textbf{ASPECT\_RATIO} in the anchor configurations remains unchanged as $[0.5,1.0,2.0]$. This hyper-parameter controls the possible changes in the aspect ratio of an anchor for object detection. 
Last, building upon the previous points, we use 12 anchors of varying sizes in each classification subnet, achieved by combining four scales and three aspect ratios. 
Finally, the classification submodule is implemented as \textit{Conv2d-ReLU-Conv2d-ReLU-Conv2d-ReLU-Conv2d-ReLU-Conv2d-Sigmoid}, where the final \textit{Sigmoid} layer generates a probability vector for classification. All convolution kernels are of size $3 \times 3$.

\textbf{Box Regression Subnet.} The box regression subnet works with the classification subnet. It generates a $4\times1$ vector to describe the four vertices of the bounding box of each detected area. %Note that the box regression subnet is architecturally identical to the classification subnet, but they do not share any parameters. 
The box regression subnet is architecturally identical to the classification subnet, except it lacks the final activation layer. Specifically, the architecture of the box regression subnet is \textit{Conv2d-ReLU-Conv2d-ReLU-Conv2d-ReLU-Conv2d-ReLU-Conv2d}. Note that the box regression subnet and the classification subnet do not share any parameters.

\subsection{Loss Function}

The total loss function used to train our proposed LithoHoD is composed of three loss terms: the  $\alpha$-balanced focal loss~\cite{lin2017focal}, the bounding box regression loss, and the distance-IoU loss~\cite{zheng2020distance}. 
The $\alpha$-balanced focal loss, used to train a classification subset, is defined as
\begin{equation}
    \mathcal{L}_\mathrm{FL} = \mathbb{E}(\ell_\mathrm{FL}) \mbox{,}
    \label{eq:focalloss1}
\end{equation}
where $\mathbb{E}(\cdot)$ denotes the expectation function, and $\ell_\mathrm{FL}$ is defined as follows: 
\begin{equation}
    \ell_\mathrm{FL}(p_{t}) = - \alpha_{t}(1-p_{t})^{\gamma}\log(p_{t}) \mbox{,}
    \label{eq:focalloss2}
\end{equation}
where $p_t$ is the probability that the training sample $x_t$ belongs to category $y$, and $\alpha$ and $\gamma$ are empirically set to be 0.25 and 2, respectively. Notice that $p_t$ is obtained by 
\begin{equation}
     p_{t} =\left\{
        \begin{array}{ll}
        p     & \mbox{, if }x_t \in y\\
        1-p   & \mbox{, otherwise}
        \end{array} \right. \mbox{.}
    \label{eq:pred_prob}
\end{equation}

Moreover, the bounding box regression loss is defined as the Huber loss (\textit{aka} smooth-L$_1$ loss). That is,
\begin{equation}
    \centering
    \mathcal{L}_{r} = \mathbb{E}(\ell_{r}) \mbox{,}
    \label{eq:bbxreg1}
\end{equation}
where 
\begin{equation}
    \centering
    \ell_{r} =\left\{
        \begin{array}{ll}
        0.5 \cdot (\mathbf{v} - \mathbf{v}_\mathrm{GT})^{2}     & \mbox{, if } |\mathbf{v} - \mathbf{v}_\mathrm{GT}|<1 \\
        |\mathbf{v} - \mathbf{v}_\mathrm{GT}|-0.5     & \mbox{, otherwise}
        \end{array} \right. \mbox{.}
    \label{eq:bbxreg2}
\end{equation}
In this equation,  $\mathbf{v}$ and $\mathbf{v}_\mathrm{GT}$ denote the predicted $4\times1$ vector specifying the bounding box and the ground-truth vector, respectively.

Finally, the the distance-IoU loss \cite{zheng2020distance} is defined as
\begin{equation}
    \mathcal{L}_\mathrm{DIoU} = \mathbb{E} (\ell_\mathrm{DIoU}) \mbox{,}
\end{equation}
and 
\begin{equation}
    \ell_\mathrm{DIoU} = 1 - \mbox{IoU} + \frac{| \mathbf{b} - \mathbf{b}_\mathrm{GT}|^2}{c_\mathrm{diag}} \mbox{.}
\end{equation}
Here, $|\mathbf{b}-\mathbf{b}_\mathrm{GT}|$ denotes the Euclidean distance between the central point of the predicted bounding box and that of the ground-truth bounding box, $c_\mathrm{diag}$ is the diagonal length of the smallest enclosing box covering the two boxes, and IoU is the \textit{Intersection-over-Union} value of the two boxes defined as 
\begin{equation}
    \mbox{IoU} = \frac{\mbox{Area of Overlap region of two Boxes}}{\mbox{Area of Union of two Boxes}}
    \mbox{.}
    \label{eq:iou}
\end{equation}

\noindent Consequently, the total loss is defined as 
\begin{equation}
    \mathcal{L}_\mathrm{total} = \mathcal{L}_{r} + \mathcal{L}_\mathrm{FL} + \lambda \mathcal{L}_\mathrm{DIoU} \mbox{,}
    \label{eq:fl}
\end{equation}
with $\lambda=0.02$.

%------------------------------------------------------------------------	
	\section{Experimental Results}
	\label{sec04:exp}
	\subsection{Datasets}

\begin{table*}[t]
\centering
\caption{Dataset details about (1) \textbf{ICCAD16} and (2) \textbf{UMC20K}. While each case in \textbf{ICCAD16} contains only one source layout, 
\textbf{UMC20K} is constructed using 1,300  process test keys} 
\footnotesize
\begin{tabular}{|c|c|cc|cc|cc|c|c|}
\hline
\textbf{Dataset} & \textbf{\# of Source Layouts} & \multicolumn{2}{c|}{\textbf{Layout size} ($\mu m^2$)} & \multicolumn{2}{c|}{\textbf{\# of Hotspots}} 
& \multicolumn{2}{c|}{\textbf{\# of Clips}} 
& \textbf{Clips size } & \textbf{Image Dimension}\\ 

&  & Train  & Test & Train & Test& Train  & Test   &  ($\mu m^2$)  & (pixels) \\ \hline
ICCAD16-2 & 1 & 6.95$\times$3.75 & 6.95$\times$3.75 & 40    & 39    & 1,000  & 8    & $2.56\times2.56$ & $256\times256$ \\
ICCAD16-3 & 1 & 12.91$\times$10.07 & 12.91$\times$10.07 & 1,388  & 1,433  & 1,000  & 33   & $2.56\times2.56$ & $256\times256$ \\
ICCAD16-4 & 1  & 79.95$\times$42.13 & $79.95\times42.13$ & 90    & 72    & 1,000  & 55   & $2.56\times2.56$ & $256\times256$ \\ \hline
%UMC20K    & 81981 & 8014  & 20120 & 2887 & $2\times2$       \\ \hline
UMC20K    & 1,300 & $200\times200$ & $200\times200$ & 81,984 & 8,011  & 20,120 & 2,887 & $2\times2$  & $512\times512$      \\ \hline
\end{tabular}
\label{table:benckmark}
\end{table*}

%\textcolor{blue}{
As detailed in Table \ref{table:benckmark}, two datasets are utilized to verify the effectiveness of the proposed framework: 1) the \textbf{ICCAD16} open dataset \cite{topaloglu2016iccad}, and 2) the \textbf{UMC20K} dataset, a private dataset provided by UMC (United Microelectronics Corporation). 
\textbf{ICCAD16} encompasses four layout designs, namely, Case-1, Case-2, Case-3, and Case-4. %Each design is shrunk to meet the design rules of the EUV (Extreme UltraViolet) metal layer. 
In each layout of  \textbf{ICCAD16}, the left-half part is used for generating training samples via random clipping, whereas the right-half part is used to generate testing samples through non-overlapping clipping. Each clipped sample is of dimension $256\times256$, corresponding to a physical size of $2.56\mu m \times 2.56 \mu m$. \textbf{ICCAD16} focuses solely on EPE (Edge Placement Error), Bridging, and Necking defects. Since Case-1 is defect-free, it is excluded from our experiments. We follow the data generation rules described in~\cite{yang2019layout,chen2019faster}.

\begin{figure}
    \centering
    \includegraphics[width=0.48\textwidth]{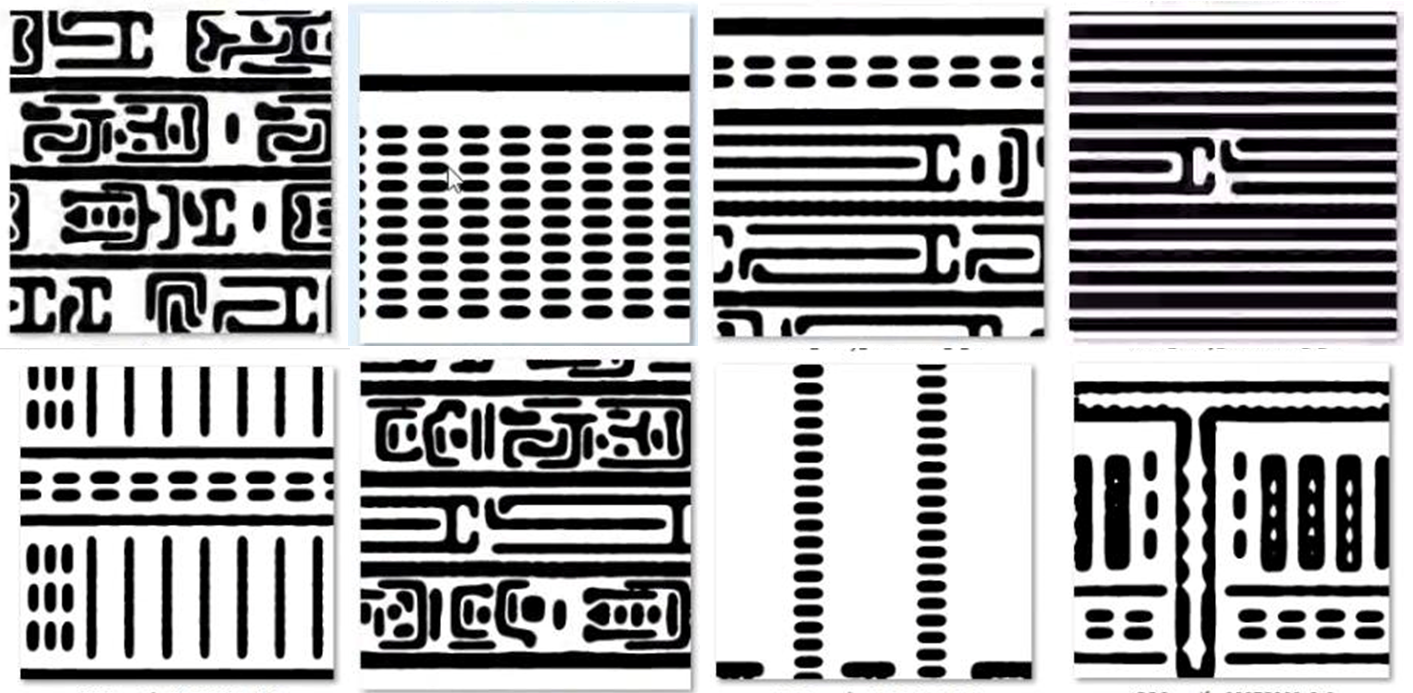}
    \caption{Example clips of \textbf{UMC20K} dataset, presented through their lithographic simulation results. The \textbf{UMC20K} dataset comprises 23,007 design clips representing digital and analog circuits, categorized into six subsets.}
    \label{fig:UMCsample}
\end{figure}

\textbf{UMC20K} comprises $20,120$ training samples and $2,887$ testing samples. 
These clips are collected from UMC's real-world process test keys, as illustrated in Fig.~\ref{fig:UMCsample}. 
Due to various functional considerations during IC circuit design, different chip regions often have different circuit appearances. For example, regions focusing on audio signal processing have appearances close to analog circuit designs, whereas regions for video signal processing exhibit digital circuit appearances. Therefore, \textbf{UMC20K} is divided into six subsets, from digital to analog circuit appearances, based on UMC's production line experience. Each sample in \textbf{UMC20K} is a $512\times512$ clip, corresponding to a physical size of $2 \mu m \times 2 \mu m$, taken from either a fabrication layout or a simulation layout. 
Additionally, each sample contains multiple hotspot areas. Since \textbf{UMC20K} is a real-world dataset that is richer and more diverse than common open datasets like \textbf{ICCAD16}, we utilize it to verify the generalizability of the proposed LithoHoD and to conduct comparisons with previous state-of-the-art methods, including BBL-HoD \cite{yang2019layout}, R-HSD \cite{chen2019faster}, and FCN-HoD \cite{gai2021flexible}.

\begin{figure}[t]
    \centering
    \includegraphics[width=0.48\textwidth]{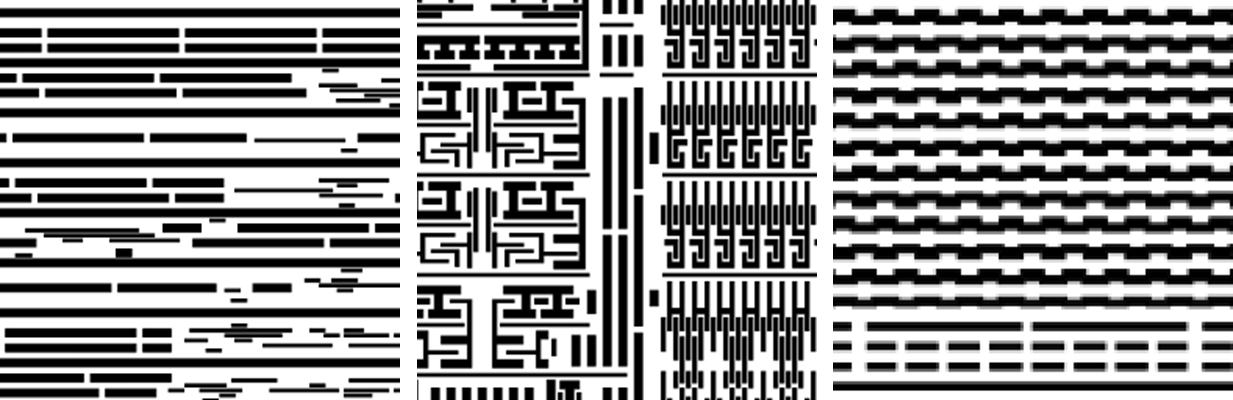}
    \caption{Examples from ICCAD16 dataset. Left: Case-2. Middle: Case-3. Right: Case-4. Each case consists of a layout sample, with its left-half used for training data and the right-half for testing data. The dimensions of the half-size for Case-2, Case-3, and Case-4 are respectively $6.95 \mu m \times 3.75 \mu m$, $12.91 \mu m \times 10.07 \mu m$, and $79.95 \mu m \times 42.13 \mu m$.    
    }  
    \label{fig:iccad_example}
\end{figure}

\subsection{Experiment Configurations}

To implement the proposed LithoHoD, we employ ResNet-34 as the feature extraction backbone used in RetinaNet. Because we utilize our LithoNet~\cite{lithonet,shao2023keeping} as the lithography simulator to guide LithoHoD, all input layout clips are resized to $512\times512$ to fit the receptive field of the pretrained LithoNet. 
The ground-truth hotspot area size of \textbf{ICCAD16} is $69\times69$, consistent with the settings used in~\cite{yang2019layout,chen2019faster}. 
For \textbf{UMC20K}, the ground-truth hotspot area size is $100\times100$.
The LithoNet model is pretrained on a dataset of paired samples to capture the deformations of circuit shapes from each layout to its corresponding SEM (Scanning Electron Microscope) image, as detailed in~\cite{lithonet}. For the subsequent experiments, we utilize this pretrained LithoNet model to implement LithoHoD and evaluate its performance on the \textbf{ICCAD16} and \textbf{UMC20K} datasets.

%Note that we 
We also set an IoU threshold for the training process. 
Specifically, a detected bounding box is classified as positive (\ie, a hotspot) if the IoU score between it and the corresponding ground-truth is greater than $0.5$. In contrast, those with an IoU score less than $0.3$ are classified as negative. 
If no positive region can be found in an input layout clip, our detector considers the one with the largest IoU value greater than $0.3$ as positive; otherwise, the clip is regarded as defect-free. 

For the experiments on \textbf{ICCAD16}, each of the three images (case-2, case-3, and case-4) in \textbf{ICCAD16} is partitioned into two halves: one half for generating randomly cropped training patches and the other for generating testing patches. 
Specifically, the experiments shown in Table~\ref{table:iccad_result} are conducted using the left half for training and the right half for testing. In contrast, Table~\ref{table:iccad_result2}(A) shows the results where the right half is used for training and the left half for testing. In the experiment shown in Table~\ref{table:iccad_result2}(B), we vertically split each image into three horizontal strips and use the middle strip for training and the bottom strip for testing since the top strip contains no hotspot.
Moreover, for the experiments of R-HSD~\cite{chen2019faster} on \textbf{ICCAD16}, by following the official setting, a detected hotspot area is considered a true positive if the distance between the predicted bounding box center and the ground-truth bounding box center is no more than $\frac{1}{3} \times 69 = 23$.  This results in an approximate IoU threshold value of %$\frac{(69-\frac{23}{\sqrt{2}})^2}{69^2\times2-(69-\frac{23}{\sqrt{2}})^2} \sim \frac{(69-16)^2}{69^2\times2-(69-16)^2}=0.4184$, 
$0.42$, 
which is more lenient than the threshold of $0.5$ used for FCN-HoD and LithoHoD.

\begin{table}[t]
\caption{Performance comparison of LithoHoD with various feature extraction backbones and three SOTA methods on \textbf{ICCAD-16}. Here, the best results are highlighted in bold, and the second-best results are underlined. Besides, the values in parentheses are the data reported in the papers of the SOTA methods, and the others are from our experimental results.}
\centering
\footnotesize
\begin{tabular}{|c|c|cccc|}
\hline
Datasets & Method & Recall & \#FA & \#FN & Time (s) \\ \hline\hline
ICCAD16  & BBL-HoD \cite{yang2019layout}     & (0.778) & (48) & (9) & (60)                 \\ %\cline{2-6}
Case-2   & R-HSD \cite{chen2019faster}     & 0.304 & 291 & 48 & 1.65 \\ 
         &      & (0.957) & (15) & (2) & (2.30) \\ %\cline{2-6}
         & FCN-HoD \cite{gai2021flexible}     & 0.058 & \textbf{21} & 65  & 22.3 \\
         &      & (0.974) & (92) & (1)  & (0.20) \\ %\cline{2-6}
         & Ours (Res18)  & \textbf{0.986} & 65 & \textbf{1} & \textbf{0.53} \\
         & Ours (Res34)  & \underline{0.971}  & \underline{55} & \underline{2} & \underline{0.55}             \\
         & Ours (Res50)  & 0.942  & 63 & $3$ & 0.58              \\
%         & Ours (Res101) & 0.913  & \textbf{37} & $4$  & 0.67       \\ 
         \hline
         
ICCAD16  & BBL-HoD \cite{yang2019layout}    & (0.912)  & (263) & (127) & (265)               \\ %\cline{2-6}
Case-3   & R-HSD \cite{chen2019faster}    & \textbf{0.983}  & 119 & \textbf{23} & 4.51 \\
         &      & (0.947)  & (78) & (76) & (10.8)  \\ %\cline{2-6}
         & FCN-HoD  \cite{gai2021flexible}   & 0.002  & \textbf{1} & 1776 & 76.04 \\
         &    & (0.978)  & (102) & (32) & (1.00) \\ %\cline{2-6}
         & Ours (Res18)  & 0.964  & 179 & 51 & \textbf{2.22} \\
         & Ours (Res34)  & 0.974  & \underline{78} & 38 & \underline{2.28}\\
         & Ours (Res50)  & \underline{0.980} & 149 & \underline{30} & 2.34 \\
%         & Ours (Res101) & 0.966  & 85 & 49 & 2.61 \\ 
         \hline
         
ICCAD16  & BBL-HoD \cite{yang2019layout}    & (1.000) & (511) & (0)   & (428.0)     \\ %\cline{2-6}
Case-4   & R-HSD  \cite{chen2019faster}   & \textbf{0.986} & \textbf{142} & \textbf{1}   & 6.35 \\
         &   & (1.000) & (92) & (0)   & (6.60) \\ %\cline{2-6}
         & FCN-HoD  \cite{gai2021flexible}   & 0 & 0 & 0 & 124.18  \\
         &      & (--)  & (--) & (--) & (--)  \\ %\cline{2-6}
         & Ours (Res18)  & \underline{0.972}  & \underline{146} & \underline{2}  & \textbf{3.32} \\
         & Ours (Res34)  & 0.944  & 170 & 4 & \underline{3.36} \\
         & Ours (Res50)  & 0.958  & 203 & 3 & 3.58   \\
%         & Ours (Res101) & \underline{0.972}  & 236 & 2  & 4.03  \\ 
         \hline
         
Average  & BBL-HoD \cite{yang2019layout}   & (0.876)  & (274)  & (45.3) & (251)                                  \\
         & R-HSD \cite{chen2019faster}    & 0.757  & 184 & 23.33 & 4.17 \\
         &      & (0.968)  & (61.6) & (26.0) & (6.50) \\
         & FCN-HoD  \cite{gai2021flexible}   & 0.020  & \textbf{7.33} & 613.67 & 74.17  \\
         &     & (--)  & (--) & (--) & (--)  \\
         & Ours (Res18)  & \textbf{0.974}  & 130  & 18.0 & \textbf{2.02} \\
         & Ours (Res34)  & \underline{0.963} & \underline{101}  & \underline{14.6} & \underline{2.06}\\
         & Ours (Res50)  & 0.960 & 138 & \textbf{12.0} & 2.16  \\
%         & Ours (Res101) & 0.950 & 119 & 18.3 & 2.43  \\ 
\hline
\end{tabular}
\label{table:iccad_result}
\end{table}

\subsection{Evaluation Metrics}

The performance of each hotspot detector is evaluated based on several metrics: the recall value (\textit{aka} true-positive-rate, TPR), the number of false alarms (\textbf{\#FA}), the number of false negative samples (\textbf{\#FN}), and runtime cost (\textbf{Time}). These metrics have been widely used in the literature \cite{gai2021flexible}. Additionally, the model performances are visualized through the TPR-to-FA (true-positive-rate to false-alarm) curve, which can be regarded as an ROC (receiver operating characteristic) curve but with an unnormalized horizontal axis. While an ROC curve plots the TPR against the false positive rate (FPR), which is commonly used to assess the performances of binary classifiers, determining the FPR in layout hotspot detection is challenging. This is because an infinite number of non-hotspot regions can be cropped from a layout, resulting in an ill-defined total false alarm count. Therefore, we replace the FPR with the false alarm count to plot the TPR-to-FA curve. Moreover, this still allows for the calculation of the AUROC (area under the ROC curve) value, as the ratio of the area under the TPR-to-FA curve to the total area remains unchanged.

%
%
\iffalse
\begin{table}[!t]
\caption{Architecture of Common ResNet Models}
\centering
\footnotesize
\begin{tabular}{|c|| c | c | c | }
\hline
 & ResNet-18 & ResNet-34 & ResNet-50 \\ 
layer name & &  &   \\ \hline
conv\_1      & \multicolumn{3}{c|}{ $7\times7$, 64, stride 2}         \\ \hline
conv\_2 & \multicolumn{3}{c|}{$3\times3$, stride 2} \\ \cline{2-4} 
        & $\times2$ & $\times3$ & $\times3$  \\ \hline
conv\_3 & $\times2$ & $\times4$ & $\times4$  \\ \hline
conv\_4 & $\times2$ & $\times6$ & $\times6$  \\ \hline
conv\_5 & $\times2$ & $\times3$ & $\times3$  \\ \hline
Residual blocks  & \multicolumn{2}{c|}{Basic block} & Bottleneck                      \\ \hline
\end{tabular}
\label{table:ResNet}
\end{table}
\fi
%
%
\begin{table}[t]
\caption{Performance comparison on \textbf{ICCAD-16}. In (A), the right half of each \textbf{ICCAD-16} sample is used to crop the training patches, and the left half is used as the testing set. In (B), the middle horizontal strip of each sample is used for training, and the bottom strip is used for testing.}
\centering
\footnotesize
\begin{tabular}{|c|c|c|cccc|}
\hline
  & Datasets & Method & Recall & \#FA & \#FN & AP \\ \hline\hline
  & ICCAD16  & R-HSD \cite{chen2019faster}     & 0.171 & 263 & 58 & 0.010\\
  & Case-2   & FCN-HoD \cite{gai2021flexible}  & 0.003 & 26 & 1011 & 0.001 \\
(A) &       & Ours (Res18)  & 0.844 & 37 & 158 & 0.706 \\ %0.7058 \\
  &         & Ours (Res34)  & 0.742 & 42 & 262 & 0.566 \\ %0.5659 \\
  &         & Ours (Res50)  & 0.514 & 9 & 493 & 0.491 \\ %0.4909 \\
         \cline{2-7}
         
&ICCAD16  &  R-HSD  \cite{chen2019faster}   & 0.971  & 176 & 74 & 0.179  \\
&Case-3   & FCN-HoD \cite{gai2021flexible}  & 0.000  & 0 & 34684 & 0.000 \\
&         & Ours (Res18)  & 0.990  & 41 & 362 & 0.928 \\%0.9276  \\
&         & Ours (Res34)  & 0.987  & 44 & 444 & 0.941 \\ %0.9413 \\
&         & Ours (Res50)  & 0.928  & 35 & 418 & 0.930 \\ %0.9302  \\
         \cline{2-7}
         
&ICCAD16  &  R-HSD  \cite{chen2019faster}   & 0.123 & 48 & 64  & 0.408  \\
&Case-4   & FCN-HoD \cite{gai2021flexible}  & 0.280  & 124 & 306 & 0.140 \\
&         & Ours (Res18)  & 0.391  & 37 & 259 & 0.359   \\
&         & Ours (Res34)  & 0.459  & 32 & 230 & 0.706 \\
&         & Ours (Res50)  & 0.435  & 35 & 240  & 0.409 \\
         \hline
  & ICCAD16  & R-HSD \cite{chen2019faster}     & 0.863 & 252 & 7 & 0.329 \\
  & Case-2   & FCN-HoD \cite{gai2021flexible}  & 0.078 & 19 & 188 & 0.043 \\
(B) &       & Ours (Res18)  & 0.882 & 19 & 24 & 0.798 \\
  &         & Ours (Res34)  & 0.804 & 25 & 40 & 0.698 \\
  &         & Ours (Res50)  & 0.726 &  7 & 56 & 0.693 \\
         \cline{2-7}
         
&ICCAD16  &  R-HSD  \cite{chen2019faster}   & 0.977  & 102 & 32 & 0.373   \\
&Case-3   & FCN-HoD \cite{gai2021flexible}  & 0.001  & 20 & 5235 & 0.000 \\
&         & Ours (Res18)  & 0.797  & 22 & 1062 & 0.797  \\
&         & Ours (Res34)  & 0.780  & 16 & 1152 & 0.779 \\
&         & Ours (Res50)  & 0.792  & 10 & 1008 & 0.791  \\
         \cline{2-7}
         
&ICCAD16  &  R-HSD  \cite{chen2019faster}   & 0.904 & 363 &  7 & 0.191   \\
&Case-4   & FCN-HoD \cite{gai2021flexible}  & 0.969  & 16 & 1 & 0.838 \\
&         & Ours (Res18)  & 0.781  &  7 &  7 & 0.673   \\
&         & Ours (Res34)  & 0.781  & 15 &  7 & 0.501 \\
&         & Ours (Res50)  & 0.781  &  5 &  7 & 0.698  \\    
\hline
\end{tabular}
\label{table:iccad_result2}
\end{table}

\begin{figure*}[!t]
     \centering
     \includegraphics[width=0.85\textwidth]{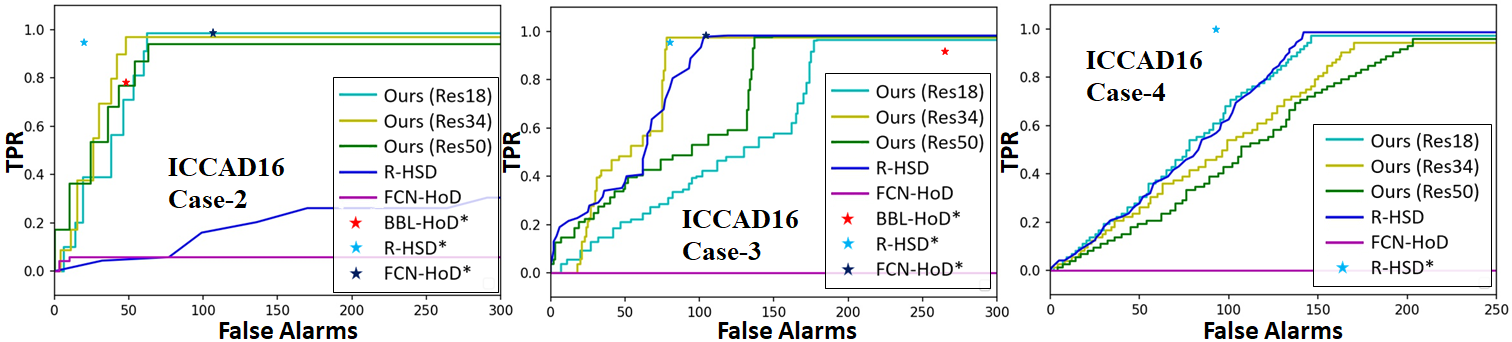}%{figure/fig06_ICCAD_comparison_new.png} %     {figure/fig06_ICCAD_comparison.png}
     \caption{Performance comparisons in terms of TPR-to-FA curves on ICCAD16 dataset. Left: Case-2; Middle: Case-3; Right: Case-4. In the figure legend, we abbreviate our LithoHoD equipped with different feature extraction backbones as Ours(Res18), Ours(Res34), and Ours(Res50). Note that BBL-HoD*, R-HSD*, and FCN-HoD* are single-point performance values reported in their papers.}
     \label{fig:iccad16}
\end{figure*}
%\iffalse
\begin{figure*}
    \centering
    \includegraphics[width=0.85\textwidth]{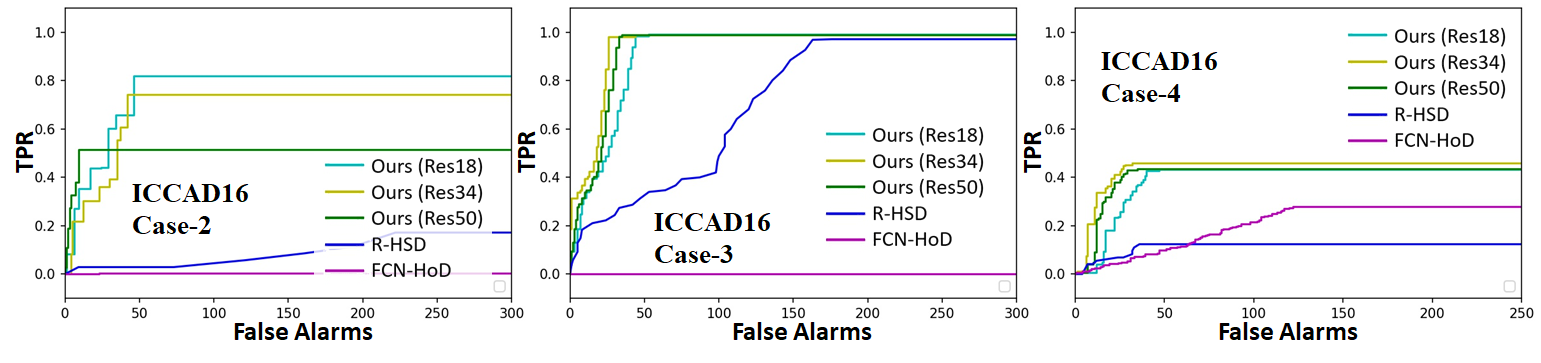}\\
    \includegraphics[width=0.85\textwidth]{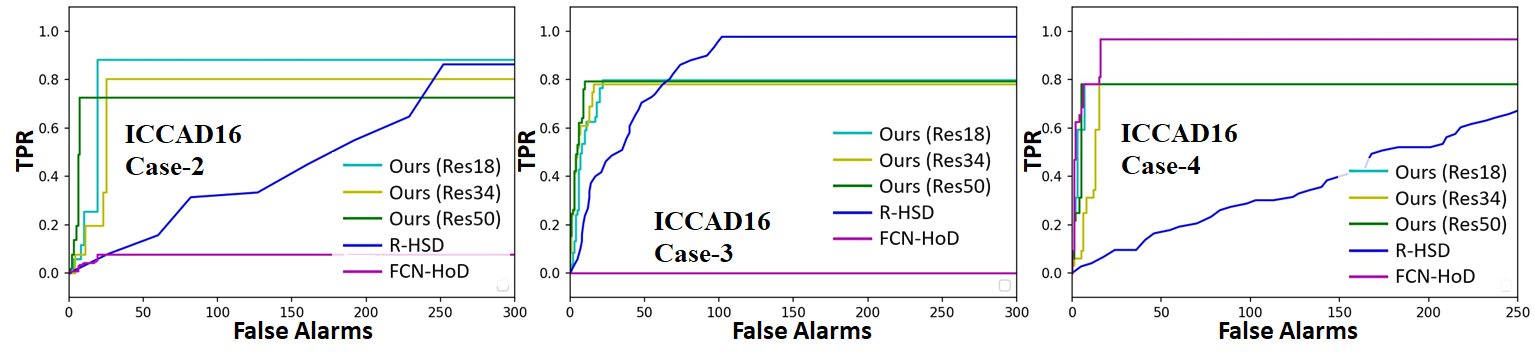}
    \caption{TPR-to-FA curves of two experiment sets on ICCAD16 dataset. From top to bottom: i) TPR-to-FA curves of the experiment set shown in Table \ref{table:iccad_result2}(A), and ii) TPR-to-FA curves of the experiment set shown in Table \ref{table:iccad_result2}(B).}
    \label{fig:rocnewexp}
\end{figure*}
%\fi

\subsection{Performance Evaluation on ICCAD16}

We compare the performance of our LithoHoD against existing state-of-the-art methods on \textbf{ICCAD16}. Since the object detection backbone of LithoHoD is a RetinaNet, whose feature extraction backbone can be readily replaced with any network in the ResNet family~\cite{he2016deep}, we evaluate the performance of LithoHoD with different feature extraction backbones, including ResNet-18, ResNet-34, and ResNet-50. 
%The architectures of these common backbones are summarized in   Table~\ref{table:ResNet}. 
Note that in Table~\ref{table:iccad_result}, the performance values in parentheses are those reported in the BBL-HoD~\cite{yang2019layout}, FCN-HoD~\cite{gai2021flexible}, and R-HSD~\cite{chen2019faster}, while the others are obtained from our experimental results. Also, as FCN-HoD and BBL-HoD do not release their official source codes, the performances with FCN-HoD are measured based on our own implementation.

First of all, Table~\ref{table:iccad_result}  shows that LithoHoD consistently outperforms the compared methods. This is attributed to LithoHoD's capability to pinpoint regions at risk of developing hotspots, guided by the pretrained LithoNet integrated into its framework. Additionally, the numbers of false negatives (\#FN) associated with LithoHoD variants are, on average, lower than those of other SOTA methods. This fact underscores the importance of incorporating a lithography simulator into the layout hotspot detection framework.

Next, Table~\ref{table:iccad_result2}(A) shows the experiments using the right half image for training and the left half for testing, and Table~\ref{table:iccad_result2}(B) shows the experiments using the middle strip for training and the bottom strip for testing. %\footnote{For Case-4, the topmost $\frac{1}{3}$ horizontal stripe area was disregarded because it contains no hotspots. Accordingly, the middle $\frac{1}{3}$ horizontal stripe area was used for training and the bottom $\frac{1}{3}$ horizontal stripe for testing.}. 
To better assess each model's robustness against data variations, we measure the average precision (AP) values denoting the area under the recall-precision curve in Table~\ref{table:iccad_result2}. A higher AP value indicates a better model capability of identifying true positives while preventing false negatives. The two experiments in Table~\ref{table:iccad_result2} show that LithoHoD outperforms the two compared SOTAs in AP in most cases, suggesting that LithoHoD can learn from and be deployed on different datasets stably.

Furthermore, experiments with our LithoHoD reveal that various feature extraction backbones are better suited to specific datasets. A deeper backbone generally leads to overfitting when the dataset is less comprehensive, as observed from \textbf{ICCAD16} Case-2. As a result, we find that LithoHoD with the ResNet-18 backbone, on average, outperforms the other variants in this case. Additionally,  Tables~\ref{table:iccad_result}~and~\ref{table:iccad_result2} show that LithoHoD with the ResNet-34 backbone achieves the most consistent performance across all metrics. Hence, we opt for this variant of LithoHoD for the ablation study and for conducting additional experiments on the real-world \textbf{UMC20K}. Finally, Fig.~\ref{fig:iccad16} and Fig.~\ref{fig:rocnewexp} demonstrate the TPR-to-FalseAlarm curves of different methods on  \textbf{ICCAD16}, further showing the superiority of LithoHoD on this open dataset.

%\textcolor{purple}{Finally, our experiments on ICCAD16 demonstrate that even under varying lithography conditions, the pretrained LithoNet can predict "relative" deformations caused by different layout patterns. The predicted deformation map can still help anticipate possible circuit shape deformations because the cross-attention mechanism will learn to appropriately weight and scale the local deformations under the supervision of hotspot ground truths. In this way, LithoHoD can still learn the relationships among deformation maps, layout patterns, and hotspot ground truths to some extent, achieving superior hotspot detection performance even under different lithography conditions. This indicates that even if the lithography condition learned by the pre-trained LithoNet and the condition learned by LithoHoD are different, the lower detector branch of LithoHoD can still learn the correspondences between the layout patterns and hotspot labels through the training process. The proposed hotspot detection framework can ignore inaccurate lithography simulation results from the upper LithoNet branch via the proposed cross-attention mechanism.  }

\begin{figure*}[!t]
    \centering
    \includegraphics[width=0.85\textwidth]{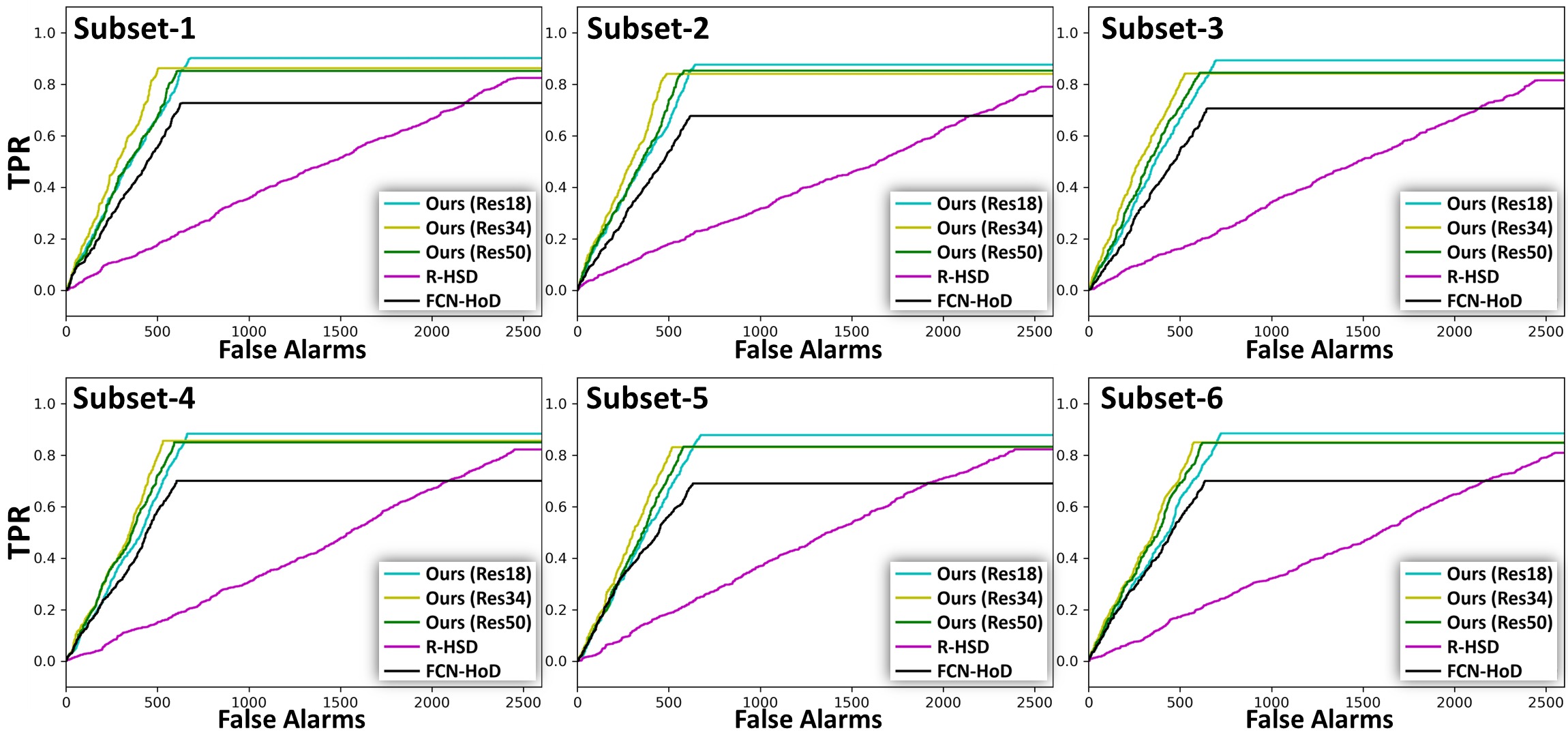}
    \caption{Comparison of TPR-to-FalseAlarm performances with various hotspot detection models on \textbf{UMC20K}. In the figure legend, we abbreviate the LithoHoD equipped with different feature extraction backbones as Ours (Res18), Ours (Res34), and Ours (Res50).  }
    \label{fig:ablation_umc}
\end{figure*}

\subsection{Performance Evaluation on UMC20K}

%\textcolor{blue}{
We also conduct a comprehensive performance evaluation on \textbf{UMC20K}, a dataset comprising a rich number of real-world layout samples. The results depicted in Fig.~\ref{fig:ablation_umc} and Table~\ref{table:umc_result} unequivocally demonstrate that LithoHoD, guided by the pretrained LithoNet, achieves the highest recall score when employing the ResNet-18 backbone.  Furthermore, LithoHoD effectively mitigates the false alarm counts, thanks to LithoNet pre-trained on real-world layout-SEM sample pairs. These findings confirm that leveraging a lithography simulator pre-trained on data collected from a similar domain can substantially enhance the efficacy of a layout hotspot detector. Note that, since the official codes for the R-HSD method~\cite{chen2019faster} and the FCN-HoD method~\cite{gai2021flexible} have not been released, all the performance values in Table~\ref{table:umc_result} and the curves in Fig.~\ref{fig:ablation_umc} are obtained from our own implementations.

\begin{table}[!t]
\caption{Performance comparison of LithoHoD with various feature extraction backbones and two SOTA methods on \textbf{UMC20K}'s testing subsets} 
\centering
\footnotesize
\begin{tabular}{|c|c|rrrr|}
\hline
Datasets & Method & Recall & \#FA & \#FN & Time (s) \\ \hline\hline
%Subset-1  & BBL-HoD \cite{yang2019layout} & aa & aa & aa & aa       \\
%Subset-1  & R-HSD \cite{chen2019faster}     & 0.808 & 2497 & 246 & 78.44 \\
Subset-1  & R-HSD \cite{chen2019faster}     & 0.824 & 2467 & 226 & 78.44 \\
    & FCN-HoD \cite{gai2021flexible}  & 0.727 & 673 & 351  & 16.01 \\
481 Images   & Ours (Res18)  & 0.901 & 683 & 127 & 32.43 \\
1284 Hotspots    %& Ours (Res34)  & 0.882 & 664 & 152 & 34.64 \\
    & Ours (Res34)  & 0.862 & 507 & 177   & 32.76   \\
    %& Ours (Res50)  & 0.870 & 608 & 167 & 34.77 \\
    & Ours (Res50)  & 0.851 & 612 & 191 & 33.13 \\
%    & Ours (Res101) & 0.896 & 699 & 134 & 36.84 \\  
    \hline
         
%Subset-2  & BBL-HoD \cite{yang2019layout} & aa & aa & aa & aa       \\
%Subset-2  & R-HSD \cite{chen2019faster}     & 0.816 & 2418 & 231 & 80.42 \\
Subset-2  & R-HSD \cite{chen2019faster}     & 0.790 & 2545 & 298 & 80.42 \\
    & FCN-HoD \cite{gai2021flexible}  & 0.678 & 619 & 458  & 15.44 \\
481 Images    & Ours (Res18)  & 0.876 & 646 & 176 & 33.83 \\
1416 Hotspots    %& Ours (Res34)  & 0.857 & 658 & 203 & 35.17 \\
    & Ours (Res34)  & 0.840 & 488 & 227   & 32.83   \\
    & Ours (Res50)  & 0.852 & 584 & 209 & 33.46 \\
%    & Ours (Res101) & 0.877 & 653 & 174 & 36.80 \\  
         \hline
         
%Subset-3  & BBL-HoD \cite{yang2019layout} & aa & aa & aa & aa       \\
%Subset-3  & R-HSD \cite{chen2019faster}     & 0.810 & 2503 & 250 & 78.84 \\
Subset-3  & R-HSD \cite{chen2019faster}     & 0.815 & 2453 & 224 & 78.84 \\
    & FCN-HoD \cite{gai2021flexible}  & 0.706 & 650 & 388  & 15.38 \\
481 Images    & Ours (Res18)  & 0.892 & 695 & 142 & 33.32 \\
1318 Hotspots    %& Ours (Res34)  & 0.864 & 657 & 179 & 34.66 \\
    & Ours (Res34)  & 0.841 & 526 & 209   & 33.51   \\
    & Ours (Res50)  & 0.845 & 607 & 205 & 33.09 \\
%    & Ours (Res101) & 0.877 & 718 & 167 & 36.54 \\  
    \hline
    
%Subset-4  & BBL-HoD \cite{yang2019layout} & aa & aa & aa & aa       \\
%Subset-4  & R-HSD \cite{chen2019faster}     & 0.809 & 2486 & 262 & 79.05 \\
Subset-4  & R-HSD \cite{chen2019faster}     & 0.822 & 2453 & 239 & 79.05 \\
    & FCN-HoD \cite{gai2021flexible}  & 0.700 & 604 & 402  & 15.30 \\
481 Images    & Ours (Res18)  & 0.883 & 663 & 157 & 33.35 \\
1339 Hotspots    %& Ours (Res34)  & 0.873 & 652 & 170 & 34.66 \\
    & Ours (Res34)  & 0.855 & 530 & 194   & 33.33   \\
    %& Ours (Res50)  & 0.857 & 607 & 192 & 35.15 \\
    & Ours (Res50)  & 0.849 & 591 & 202 & 33.04 \\
%    & Ours (Res101) & 0.887 & 720 & 152 & 36.35 \\ 
    \hline
         
%Subset-5  & BBL-HoD \cite{yang2019layout} & aa & aa & aa & aa       \\
%Subset-5  & R-HSD \cite{chen2019faster}     & 0.811 & 2415 & 241 & 78.94 \\
Subset-5  & R-HSD \cite{chen2019faster}     & 0.822 & 2397 & 227 & 78.94 \\
    & FCN-HoD \cite{gai2021flexible}  & 0.699 & 632 & 396  & 15.23 \\
481 Images    & Ours (Res18)  & 0.878 & 675 & 156 & 32.35 \\
1276 Hotspots    %& Ours (Res34)  & 0.859 & 665 & 180 & 35.12 \\
    & Ours (Res34)  & 0.831 & 519 & 216   & 32.89   \\
    %& Ours (Res50)  & 0.848 & 591 & 194 & 34.43 \\
    & Ours (Res50)  & 0.832 & 580 & 214 & 33.09 \\
%    & Ours (Res101) & 0.868 & 702 & 169 & 36.32 \\ 
\hline
         
%Subset-6  & BBL-HoD \cite{yang2019layout} & aa & aa & aa & aa       \\
%Subset-6  & R-HSD \cite{chen2019faster}     & 0.821 & 2486 & 247 & 78.64 \\
Subset-6  & R-HSD \cite{chen2019faster}     & 0.808 & 2550 & 264 & 78.64 \\
    & FCN-HoD \cite{gai2021flexible}  & 0.700 & 636 & 414  & 15.41 \\
482 Images    & Ours (Res18)  & 0.844 & 726 & 160 & 34.52 \\
1378 Hotspots    %& Ours (Res34)  & 0.867 & 653 & 184 & 35.17 \\
    & Ours (Res34)  & 0.849 & 578 & 208   & 33.09   \\
    %& Ours (Res50)  & 0.858 & 639 & 196 & 35.75 \\
    & Ours (Res50)  & 0.847 & 625 & 211 & 33.49 \\
%    & Ours (Res101) & 0.887 & 753 & 156 & 36.27 \\ 
\hline
         
%Average  & BBL-HoD \cite{yang2019layout} & aa & aa & aa & aa         \\
Average  & R-HSD \cite{chen2019faster}     & 0.813 & 2478 & 246 & 79.06 \\
    & FCN-HoD \cite{gai2021flexible}  & 0.701 & 630 & 402  & 15.46 \\
%Total & Ours (Res18)  & 0.890 & 777 & 147 & 35.35 \\
%8011 Hotspots %& Ours (Res34)  & 0.867 & 655 & 178 & 34.90 \\
%    & \textcolor{gray}{New (Res34)}  & \textcolor{gray}{0.846} & \textcolor{gray}{525} & \textcolor{gray}{--}   & \textcolor{gray}{34.90}   \\
%    & Ours (Res50)  & 0.852 & 609 & 207 & 34.98 \\
% %   & Ours (Res101) & 0.882 & 708 & 159 & 36.52 \\
Total & Ours (Res18)  & 0.889 & 681 & 153 & 33.33 \\
8011 Hotspots 
    & Ours (Res34)  & 0.846 & 525 & 205   & 33.06   \\
    & Ours (Res50)  & 0.846 & 600 & 205 & 33.22  \\
 \hline
\end{tabular}
\label{table:umc_result}
\end{table}

\begin{figure*}[!t]
     \centering
     \includegraphics[width=0.85\textwidth]{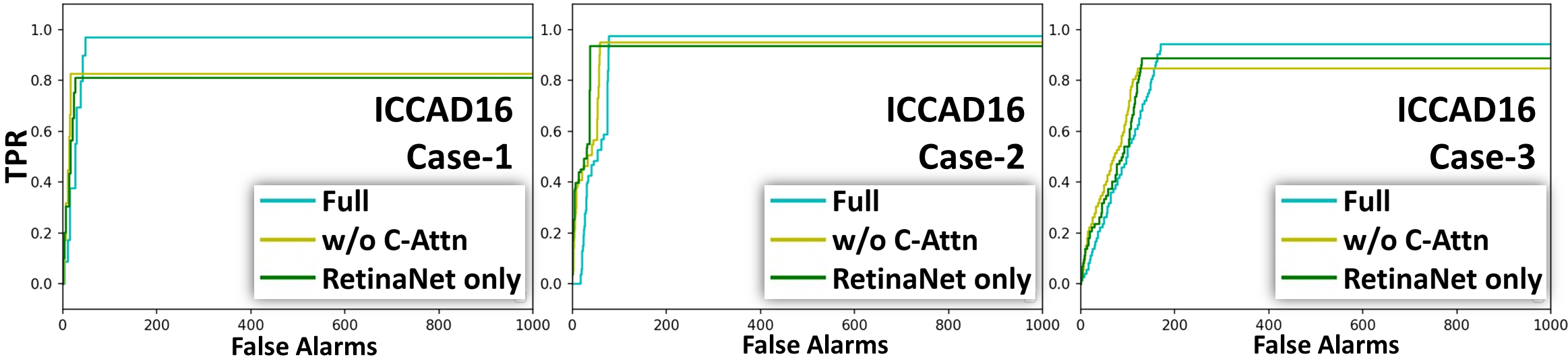}
     \caption{Ablation Study of LithoHoD on \textbf{ICCAD16} in terms of Recall to False-alarm curves. Left: Case-2, Middle: Case-3, Right: Case-4.}
     \label{fig:ablation}
\end{figure*}

Furthermore, our LithoHoD consistently outperforms previous state-of-the-art methods under the guidance of the lithography simulator, regardless of the type of feature extractor (\eg, ResNet) used. However, the experimental results suggest that LithoHoD with shallower feature extractors (\eg~ResNet-18) generally outperforms its counterparts with deeper feature extractors on \textbf{UMC-20K}. We attribute this to the fact that IC layout designs typically consist of binary polygons, involving primarily straight lines and rectangles in black-and-white format. Therefore, employing a shallower and simpler feature extractor with fewer parameters often yields superior results. In contrast, deeper feature extraction networks performing well on complex natural images (\eg, street scenes, remote sensing, and product recognition) are less effective in the context of layout hotspot detection. Consequently, this enables our LithoHoD equipped with ResNet-18 to achieve better and more stable hotspot detection results on real-world data.

Experiments on \textbf{ICCAD16} and \textbf{UMC20K} demonstrate the efficacy of LithoHoD in two distinct scenarios. The first scenario, corresponding to experiments on \textbf{UMC20K}, is that the lithography conditions used for hotspot detection are consistent with the lithography conditions of LithoNet's training dataset. The experiments on \textbf{UMC20K} show that when the lithography conditions are the same or similar, the deformation map predicted by LithoNet aids the hotspot detector well, leading to superior detection performance. % with the lithography conditions of LithoNet's training dataset. 

On the other hand, corresponding to experiments on \textbf{ICCAD16}, the second scenario is that there exists a significant domain gap between the lithography conditions for the pre-trained LithoNet and the hotspot detector. In this scenario, the deformation map predicted by LithoNet will be less accurate. However, even with inconsistent lithography conditions,  LithoNet can still predict ``relative'' local deformations caused by different layout patterns. With the supervision of the hotspot labels, these local deformation features can be properly scaled and weighted through the self- and cross-attention mechanisms of the CMF module. In this way, LithoHoD can tolerate the inaccuracies of deformation features to some extent by learning to select useful deformation features during feature fusion via its self- and cross-attention mechanisms. This is evidenced by the ablation study on \textbf{ICCAD16} (see Table~\ref{table:ablation}), where LithoNet was pre-trained on the UMC dataset.

In the worst-case scenario, where the domain gap in lithography conditions is so large that LithoNet becomes ineffective, LithoHoD will revert to functioning as a traditional hotspot detector. In this situation, its attention mechanisms will discard LithoNet's deformation features.

In summary, the experimental results support that our LithoHod is effective even without fine-tuning its lithography simulator. It can still learn the relationship between the deformation maps, layouts, and hotspot ground truths through the training routine, thereby improving hotspot detection performance, as evidenced by the results on \textbf{ICCAD16}. Nevertheless, if LithoNet can be fine-tuned based on the current lithography conditions to bridge the domain gap, the performance gain with LithoHoD will be significant, as demonstrated by the results on \textbf{UMC20K}.

\subsection{Ablation Studies}

We conduct two ablation studies to validate the individual contributions of LithoHod's modules to the performance of LithoHoD on \textbf{ICCAD16} and \textbf{UMC20K}, respectively.  Fig.~\ref{fig:ablation} and Table~\ref{table:ablation} present the results of the ablation study on \textbf{ICCAD16}. This ablation study verifies how individual modules contribute to LithoHoD by individually removing the two modules: i) the lithography simulator 
and ii) the channel-attention module within the object detection backbone. We use the term ``RetinaNet only'' to denote the first scenario in which the pretrained LithoNet is removed by deactivating the cross-model feature fusion (CMF) module. 
Second, ``w/o C-Attn'' indicates the removal of the channel-wise attention module in between ResNet and FPN, \ie, the C-Attn module shown in Fig. \ref{fig:proposed_arch}, implying the use of an unmodified RetinaNet backbone.

\begin{figure*}[!t]
     \centering
     \includegraphics[width=0.96\textwidth]{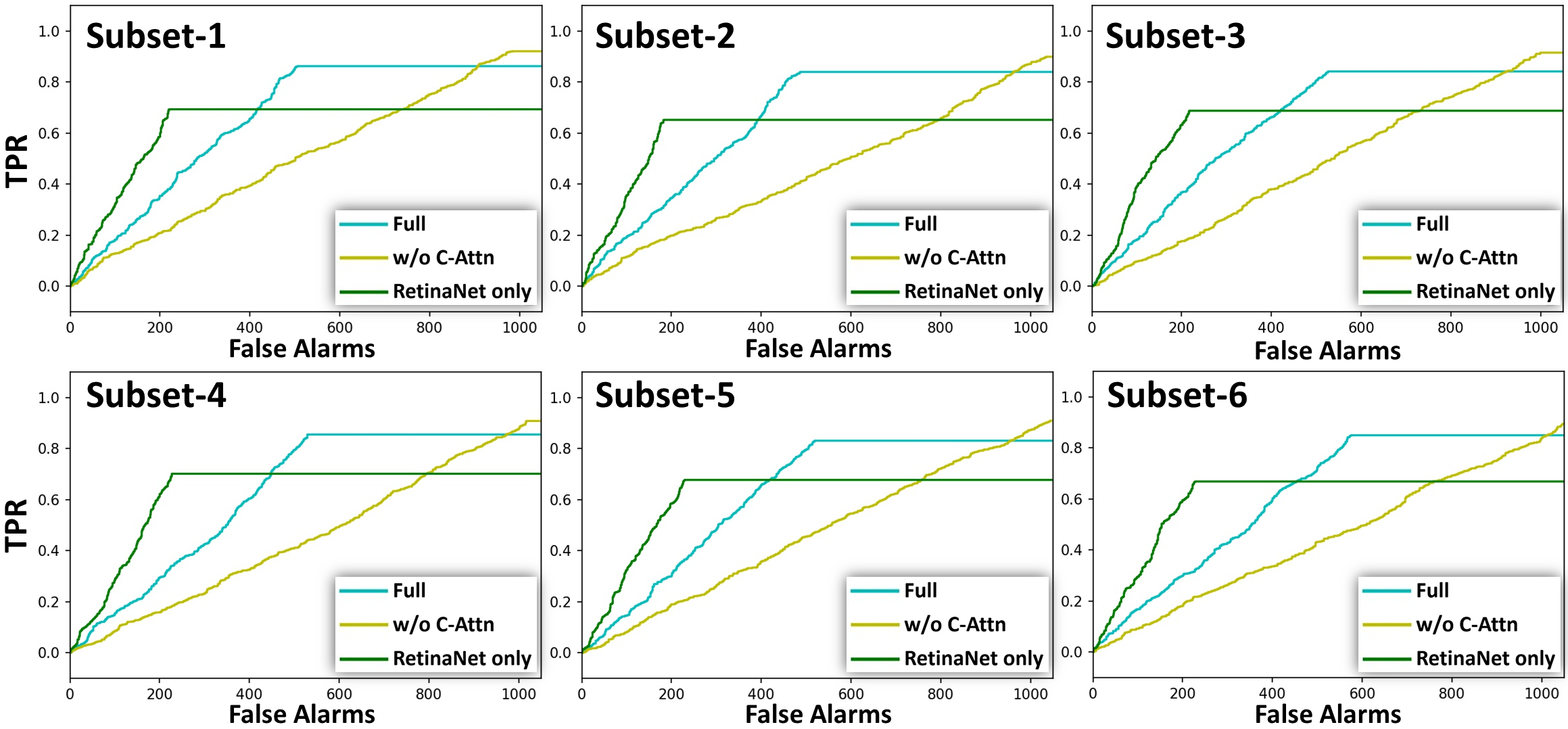}
     \caption{Ablation Study of LithoHoD on \textbf{UMC20} in terms of Recall to False-alarm curves. }
     \label{fig:UMCablationROC}
\end{figure*}
\begin{table}[!t]
\centering
\caption{Ablation study of the modules of LithoHoD on \textbf{ICCAD16}, where the best AUC socres are highlighted in bold}
\footnotesize
\begin{tabular}{|l|lc|ccc|}
\hline
Benchmark & \multicolumn{2}{c|}{Method} & Recall  & \#FA & AUC score  \\ \hline\hline

ICCAD16   & (a) & RetinaNet only & $0.812$ & $25$  & $0.800$  \\
Case-2    & (b) & w/o C-Attn & $0.826$ & $17$ & $0.818$  \\
          & (c) & Full Model   & $0.971$ & $55$  & $\textbf{0.946}$ \\
%          & (d) & Full Model   & $\textbf{0.971}$ & $55$  & $\textbf{0.946}$ \\ 
\hline
          
ICCAD16   & (a) & RetinaNet only & 0.958 & 60 & 0.916 \\
Case-3    & (b) & w/o C-Attn & 0.949 & 60 & 0.921 \\
          & (c) & Full Model   &  0.974 & 78  & \textbf{0.925} \\
%          & (d) & Full Model   &  \textbf{0.974} & 78  & \textbf{0.925} \\ 
          \hline
          
ICCAD16   & (a) & RetinaNet only & $0.903$ & 130 & 0.827 \\
Case-4    & (c) & w/o C-Attn & $0.847$ & 121 & 0.798  \\
          & (c) & Full Model & $0.944$ & 170 & $\textbf{0.860}$ \\
%          & (d) & Full Model & $\textbf{0.944}$ & 170 & $\textbf{0.860}$ \\ 
          \hline
\end{tabular}
\label{table:ablation}
\end{table}
\begin{table}[!t]
\centering
\caption{Ablation study on the modules of LithoHoD (with Res34 backbone) on \textbf{UMC20K}, where the best results are highlighted in bold}
\footnotesize
\begin{tabular}{|l|lc|ccc|}
\hline
Benchmark & \multicolumn{2}{c|}{Method} & Recall  & \#FA & AUC   \\ \hline\hline

Subset-1   & (a) & RetinaNet only & 0.693 & 222  & 0.621  \\
481 Images     & (b)  & w/o C-Attn & 0.921 & 985 & 0.504  \\
1284 Hotspots  & (c) & Full Model   & 0.862 & 507 & \textbf{0.656} \\ \hline
 %         & (d) & Full Model   & 0.862 & 507 & 0.656 \\ \hline
          
Subset-2   & (a) & RetinaNet only & 0.652 & 183  & 0.539  \\
481 Images    & (b) & w/o C-Attn & 0.899 & 1036 & 0.450  \\
1416 Hotspots & (c) & Full Model   & 0.840 & 488 & \textbf{0.644} \\ \hline
%          & (d) & Full Model   & 0.840 & 488 & 0.644 \\ \hline
          
Subset-3   & (a) & RetinaNet only & 0.687 & 217  & 0.620 \\
481 Images & (b) & w/o C-Attn & 0.915 & 1019 & 0.449  \\
1318 Hotspots & (c) & Full Model   & 0.841 & 526 & \textbf{0.644} \\ \hline
%          & (d) & Full Model   & 0.841 & 526 & 0.644 \\ \hline

Subset-4   & (a) & RetinaNet only & 0.701 & 227  & 0.621 \\
481 Images & (b) & w/o C-Attn & 0.908 & 1019 & 0.449 \\ 
1339 Hotspots & (c) & Full Model   & 0.855 & 530 & \textbf{0.625} \\ \hline
%          & (d) & Full Model   & 0.855 & 530 & 0.625 \\ \hline

Subset-5   & (a) & RetinaNet only & 0.677 & 229  & 0.604  \\
481 Images    & (b) & w/o C-Attn & 0.909 & 1043 & 0.466  \\
1276 Hotspots & (c) & Full Model   & 0.831 & 519 & \textbf{0.625} \\ \hline
%          & (d) & Full Model   & 0.831 & 519 & 0.625 \\ \hline

Subset-6   & (a) & RetinaNet only & 0.668 & 229  & 0.598  \\
482 Images & (b) & w/o C-Attn & 0.895 & 1049 & 0.447  \\
1378 Hotspots & (c) & Full Model   & 0.849 & 578 & \textbf{0.615} \\ \hline
%          & (d) & Full Model   & 0.849 & 578 & 0.615 \\ \hline          
\end{tabular}
\label{table:ablation_umc}
\end{table}

Table~\ref{table:ablation} demonstrates a significant drop in recall and AUC scores when deactivating any of these modules. Because the results in Table~\ref{table:ablation} are obtained with the models trained on \textbf{ICCAD16}, the increase in false alarms can be attributed to two factors. The first factor is the similarities among patterns in \textbf{ICCAD16}'s three cases. The second is that, by leveraging the lithography simulator, LithoHoD can detect those areas in \textbf{ICCAD16} that are not labeled as hotspots but have the potential to develop hotspots, thereby potentially leading to more false alarms.

Table~\ref{table:ablation_umc} and Fig.~\ref{fig:UMCablationROC} show the ablation study results on \textbf{UMC-20K}, demonstrating the individual contributions of LithoHoD's modules on real-world hotspot detection cases. Similar to the findings reported in Table~\ref{table:ablation}, the removal of individual modules leads to an AUC performance drop, highlighting the indispensability of each module in enhancing the model performance.
Besides, while the full model equipped with a lithography simulator tends to generate a higher number of false alarms (\#FA) compared to the object detector alone (\ie,  ``RetinaNet only''), the superior recall and AUC scores of our full model still suggest that a lithography simulator can facilitate hotspot detection. This demonstrates providing the object detector with a predicted lithography deformation map effectively aids in identifying potential hotspot areas in real-world cases.
Furthermore, the results in ``w/o C-Attn'' demonstrate a significant increase in false alarms and low AUC scores with illusory high recall values when the channel-attention module is removed. This suggests that the channel-attention module in our object detection backbone effectively reduces false alarms, albeit slightly decreasing the recall value, thus contributing to an optimized full model in terms of AUC. 

The above ablation studies reveal that integrating the lithography deformation features with the layout shape features can more effectively identify layout hotspot areas, including potentially risky regions, in real-world scenarios.

%------------------------------------------------------------------------	
	\section{Conclusion}
	\label{sec:conclusion}

In this work, we proposed a lithography simulator-guided hotspot detection framework, namely LithoHoD, capable of integrating lithography deformation features with the object detection features of a given layout. The proposed LithoHoD mainly consists of two modules:  a shape feature extractor (RetinaNet) and a lithography simulator (LithoNet). Additionally, we introduced a cross-domain attention mechanism to fuse the shape features from RetinaNet with the local lithography deformation features from LithoNet. Our experimental results confirm that the incorporation of a lithographic simulator significantly enhances the adaptability of a conventional object detection model for layout hotspot detection tasks.  Notably, our LithoHoD demonstrates superior performances on both the simulated ICCAD16 dataset and a real-world UMC20K dataset provided by an IC-fab foundry. While this simulator-guided framework may entail a slightly higher computational time complexity than previous state-of-the-art methods, its recall value and AUC score outperform these methods, setting a new benchmark in hotspot detection for IC fabrication. This work demonstrates the successful repurposing of object detectors for hotspot detection tasks with appropriately learned features, thereby laying a foundation for the future development of more advanced hotspot detection models.

%\textcolor{blue}{Finally, the foundries and hotspots presented in this paper are provided by a semiconductor manufacturer, with labeled hotspots indicating circuit patterns that can cause defects and directly affect manufacturing yield. Therefore, our current method design is unrelated to layout parasitics and does not concern power, performance, or area (PPA). One possible future extension is to redefine hotspots to reflect real PPA impacts, enabling the hotspot detection model to integrate with new EDA tools that introduce mitigation strategies.}

	
	
	\bibliographystyle{IEEEtran}
	\bibliography{LithoDet_references}

	\vspace{-0.4in}
\begin{IEEEbiography}
[{\includegraphics[width=1in,height=1.25in,clip,keepaspectratio]{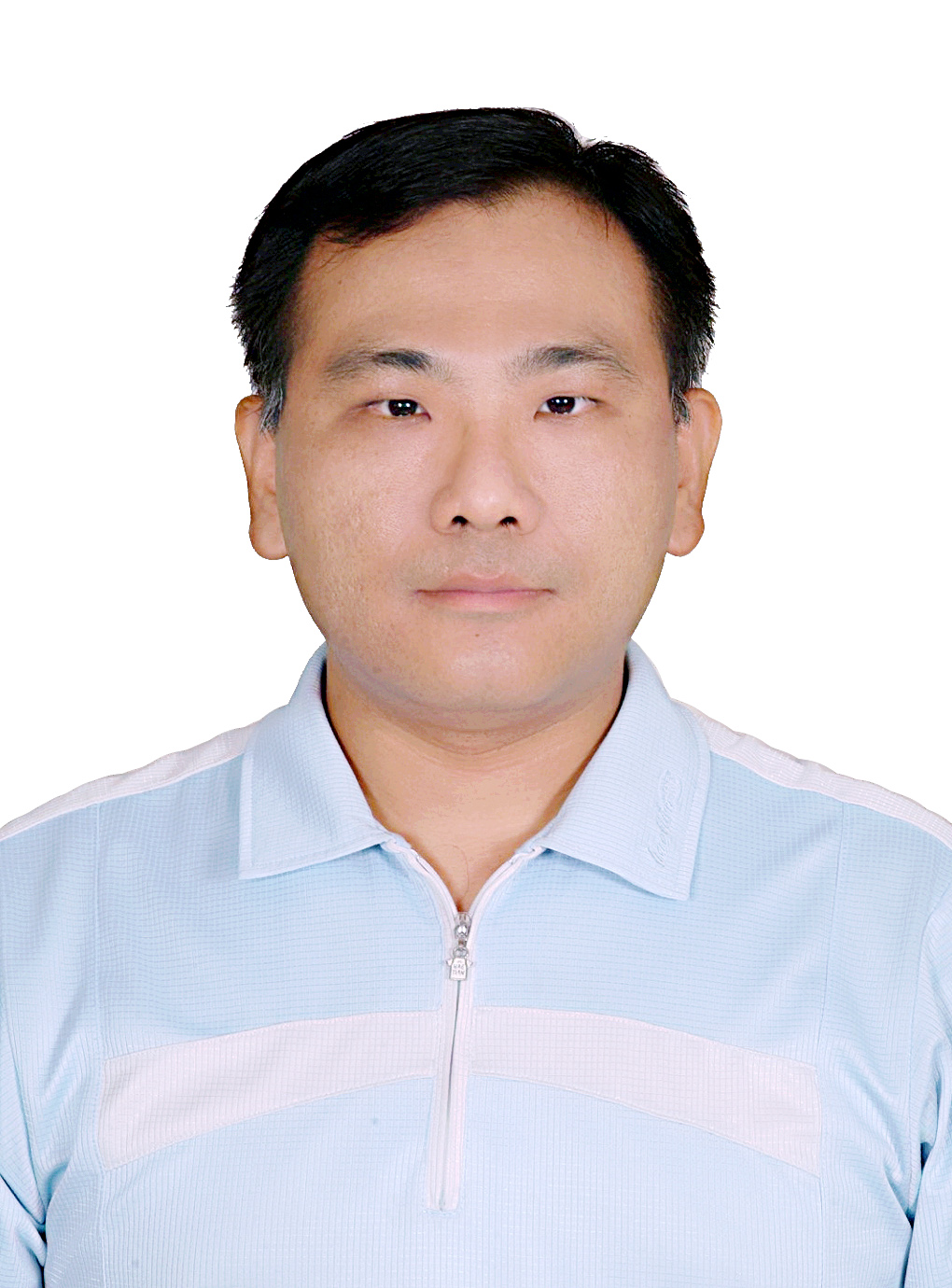}}]
{Hao-Chiang~Shao}
(Member, IEEE) received his Ph.D. degree in electrical engineering from National Tsing Hua University, Taiwan, in 2012. 
He has been an Assistant Professor with the Inst. Data Science and Information Computing, National Chung Hsing University, Taiwan, since 2022. During 2018 to 2022, he was an Assistant Professor with the Dept. Statistics and Information Science, Fu Jen Catholic University, Taiwan. In 2017--2018, he was an R\&D engineer with the Computational Intelligence Technology Center, Industrial Technology Research Institute, Taiwan, taking charges of DNN-based automated optical inspection projects; during 2012 to 2017, he was a postdoctoral researcher with the Institute of Information Science, Academia Sinica, involved in \textit{Drosophila} brain research projects. His research interests include 2D+Z image atlasing and 3D mesh processing, big image data analysis, classification methods, deep learning and computer vision, and image forgery detection techniques.
\end{IEEEbiography}

%\newpage

\vspace{-0.4in}
\begin{IEEEbiography}
	[{\includegraphics[width=1in,height=1.25in,clip,keepaspectratio] {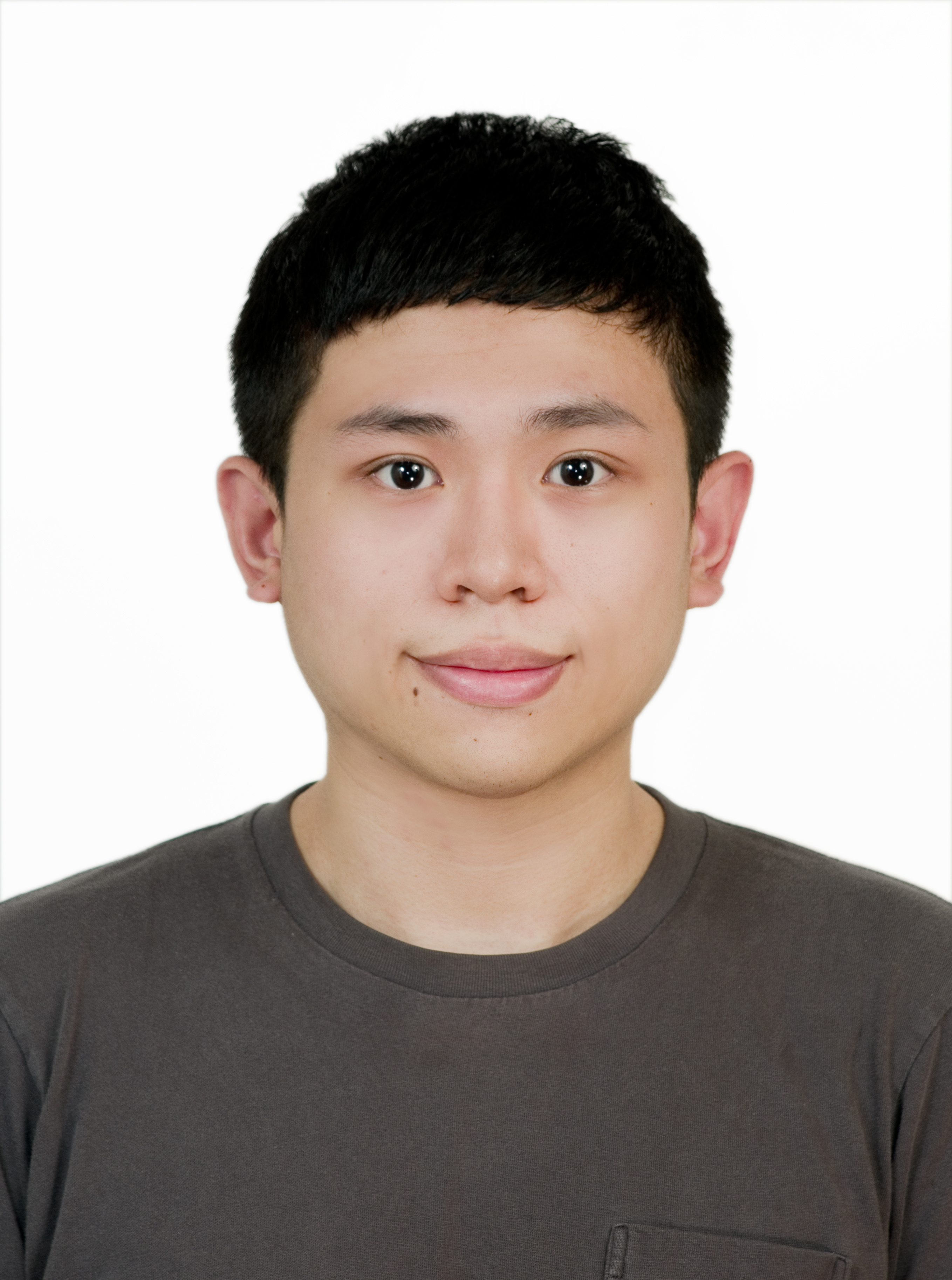}}]
	{Guan-Yu Chen}
	 received his B.S. degree in Electronic Engineering from National Central University in 2020 and M.S. degree in Electrical Engineering from National Tsing Hua University in 2023.  
	He joined Novatek Microelectronics Corp., Hsinchu, Taiwan, as a software/firmware engineer in 2023.   His research interests lie in computer vision, machine learning, and visual analytics for IC design for manufacturability.
\end{IEEEbiography}

\vspace{-0.4in}
\begin{IEEEbiography}
	[{\includegraphics[width=1in,height=1.25in,clip,keepaspectratio] {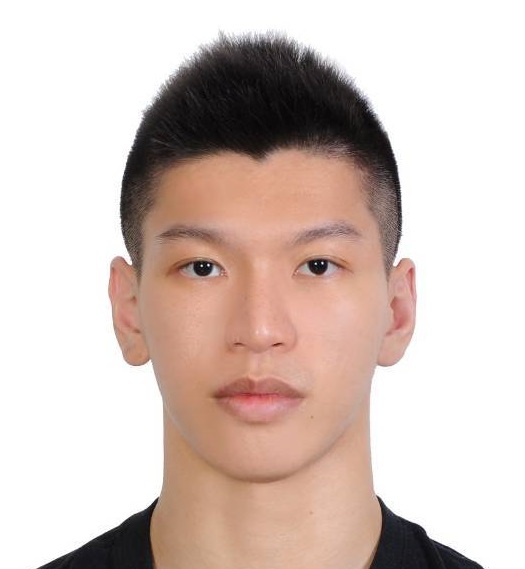}}]
	{Yu-Hsien Lin} received the B.S. degree in computer science from the Fu Jen Catholic University, New Taipei City, Taiwan, in 2020, Taiwan, the M.A. degree in applied statistics from Fu Jen Catholic University, in 2022. He is currently pursuing his Ph.D. degree at the Department of Electrical Engineering of National Tsing Hua University, Hsinchu, Taiwan.
His research interests mainly lie in machine learning, electronic design automation, and computer vision. 
\end{IEEEbiography}

\vspace{-0.4in}
\begin{IEEEbiography}
	[{\includegraphics[width=1in,height=1.25in,clip,keepaspectratio] {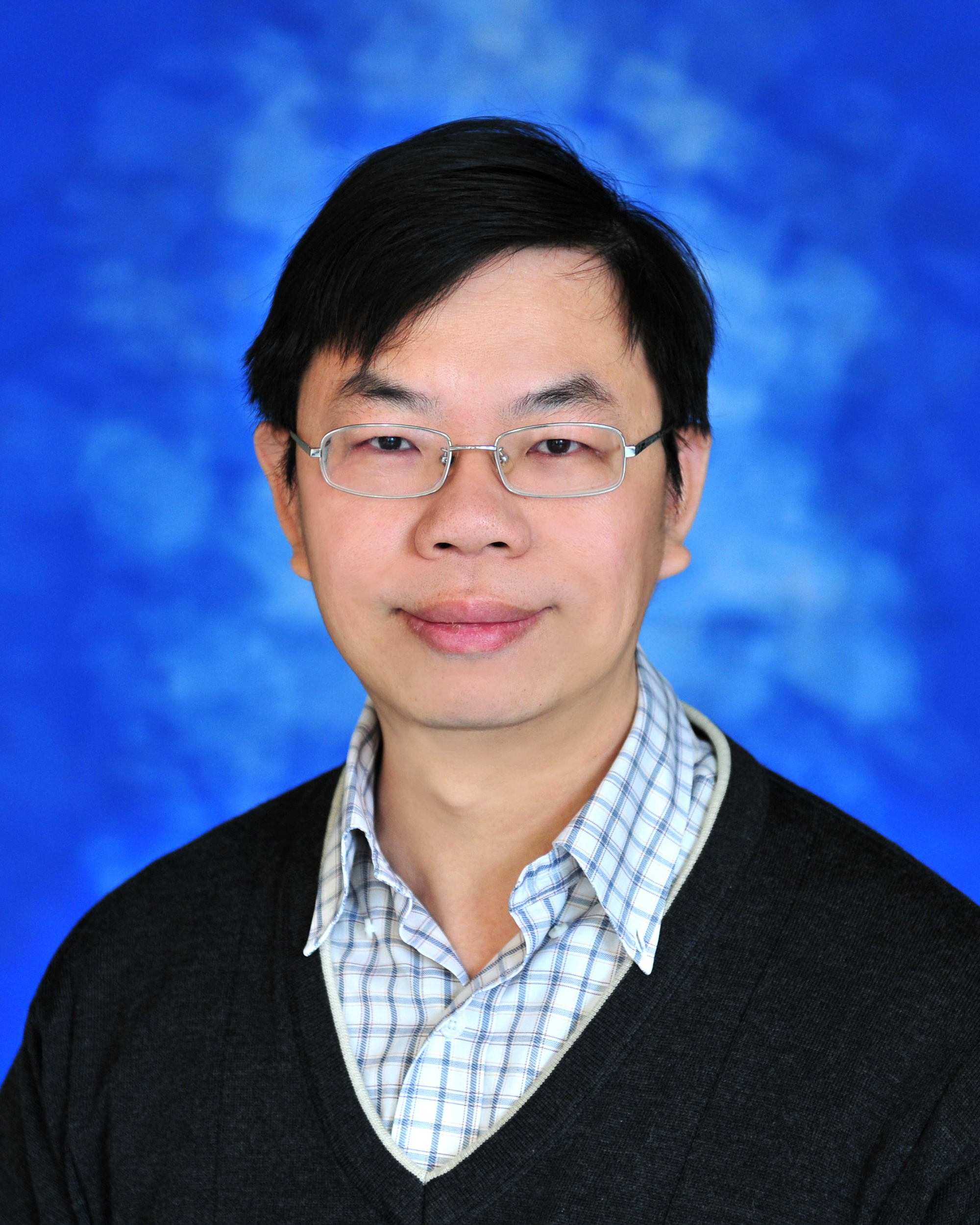}}]
	{Chia-Wen Lin}
	(Fellow, IEEE) received his Ph.D. degree from National Tsing Hua University (NTHU), Hsinchu, Taiwan, in 2000.  
	Dr. Lin is currently a Distinguished Professor with the Department of Electrical Engineering and the Institute of Communications Engineering, NTHU. His research interests include image/video processing and computer vision.  He has served as a Fellow Evaluation Committee member (2021--2023), BoG Members-at-Large (2022--2024), and Distinguished Lecturer (2018--2019) of IEEE CASS.   He was Chair of IEEE ICME Steering Committee (2020--2021). He served as TPC Co-Chair of IEEE ICIP 2019 and IEEE ICME 2010, and General Co-Chair of IEEE VCIP 2018.  He received two best paper awards from VCIP 2010 and 2015. He was an Associate Editor of \textsc{IEEE Transactions on Image Processing}, \textsc{IEEE Transactions on Circuits and Systems for Video Technology}, \textsc{IEEE Transactions on Multimedia}, and \textsc{IEEE Multimedia}. 
\end{IEEEbiography}

\vspace{-0.4in}
\begin{IEEEbiography}
	[{\includegraphics[width=1in,height=1.25in,clip,keepaspectratio] {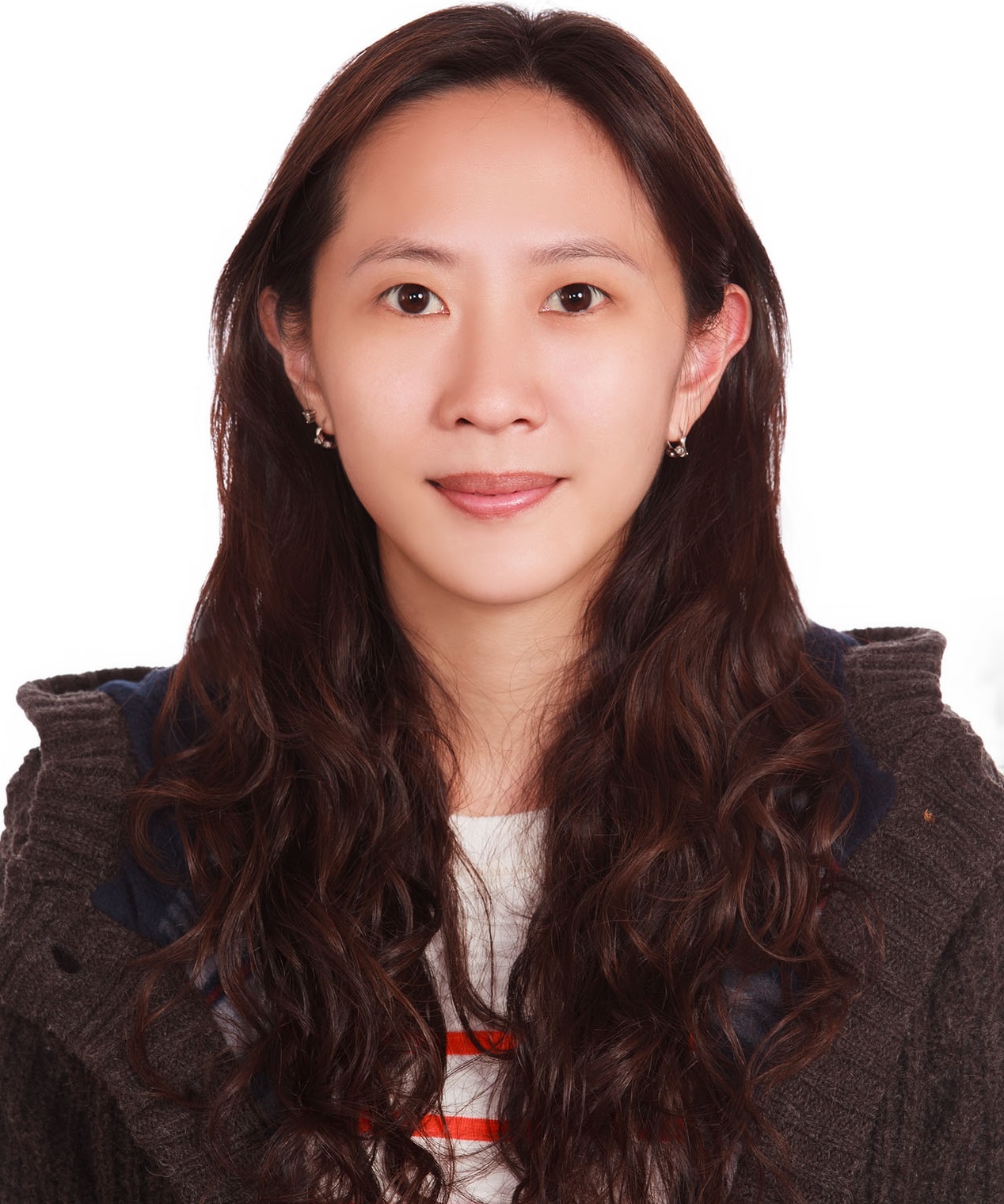}}]
	{Shao-Yun Fang}
	(Member, IEEE) received a B.S. degree in electrical engineering from the National Taiwan University (NTU), Taipei, Taiwan, in 2008 and a Ph.D. degree from the Graduate Institute of Electronics Engineering, NTU, in 2013. She is currently a Professor of the Department of Electrical Engineering, National Taiwan University of Science and Technology, Taipei, Taiwan. Her research interests focus on physical design and design for manufacturability for integrated circuits. Dr. Fang received two Best Paper Awards from the 2016 International Conference on Computer Design and the 2016 International Symposium on VLSI Design, Automation, and Test, and two Best Paper Nominations from  International Symposium on Physical Design in 2012 and 2013.
\end{IEEEbiography}

\vspace{-0.4in}
\begin{IEEEbiography}
	[{\includegraphics[width=1in,height=1.25in,clip,keepaspectratio] {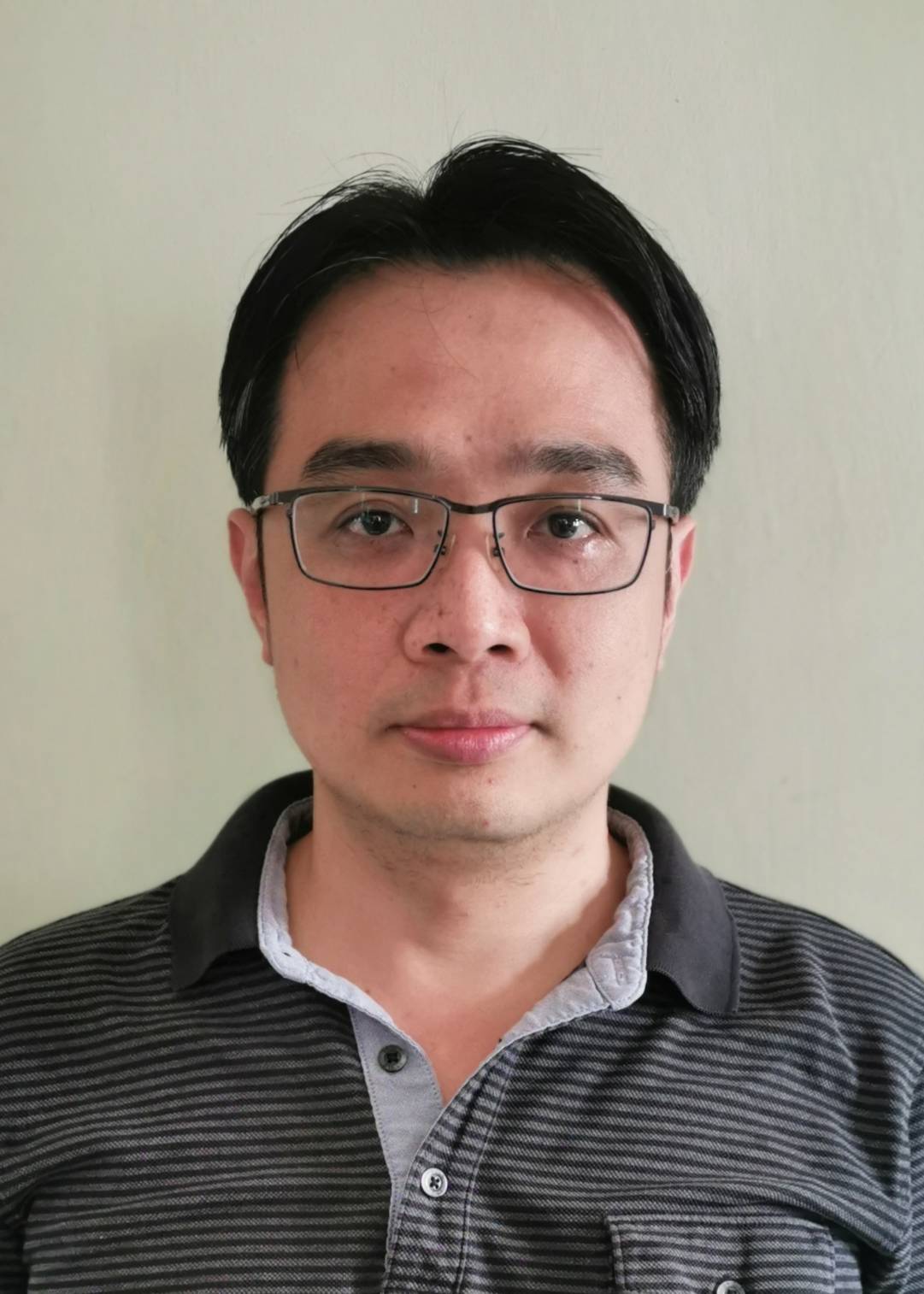}}]
	{Pin-Yian Tsai}
    received his M.S. degree in Physics from National Tsing Hua University (NTHU), Taiwan, in 2008. He is currently a technical manager of the Product Engineering Department in United Microelectronics Corporation (UMC). He led the launch of UMC’s first 14nm product tape out (2017) and is currently working and researching on the field of Design for Manufacturing (DFM). He is now focusing on developing methods for predicting weak patterns in layout manufacturing and automatic optical proximity correction (OPC) to improve the manufacturing yield.
\end{IEEEbiography}

\vspace{-0.4in}
\begin{IEEEbiography}
	[{\includegraphics[width=1in,height=1.25in,clip,keepaspectratio] {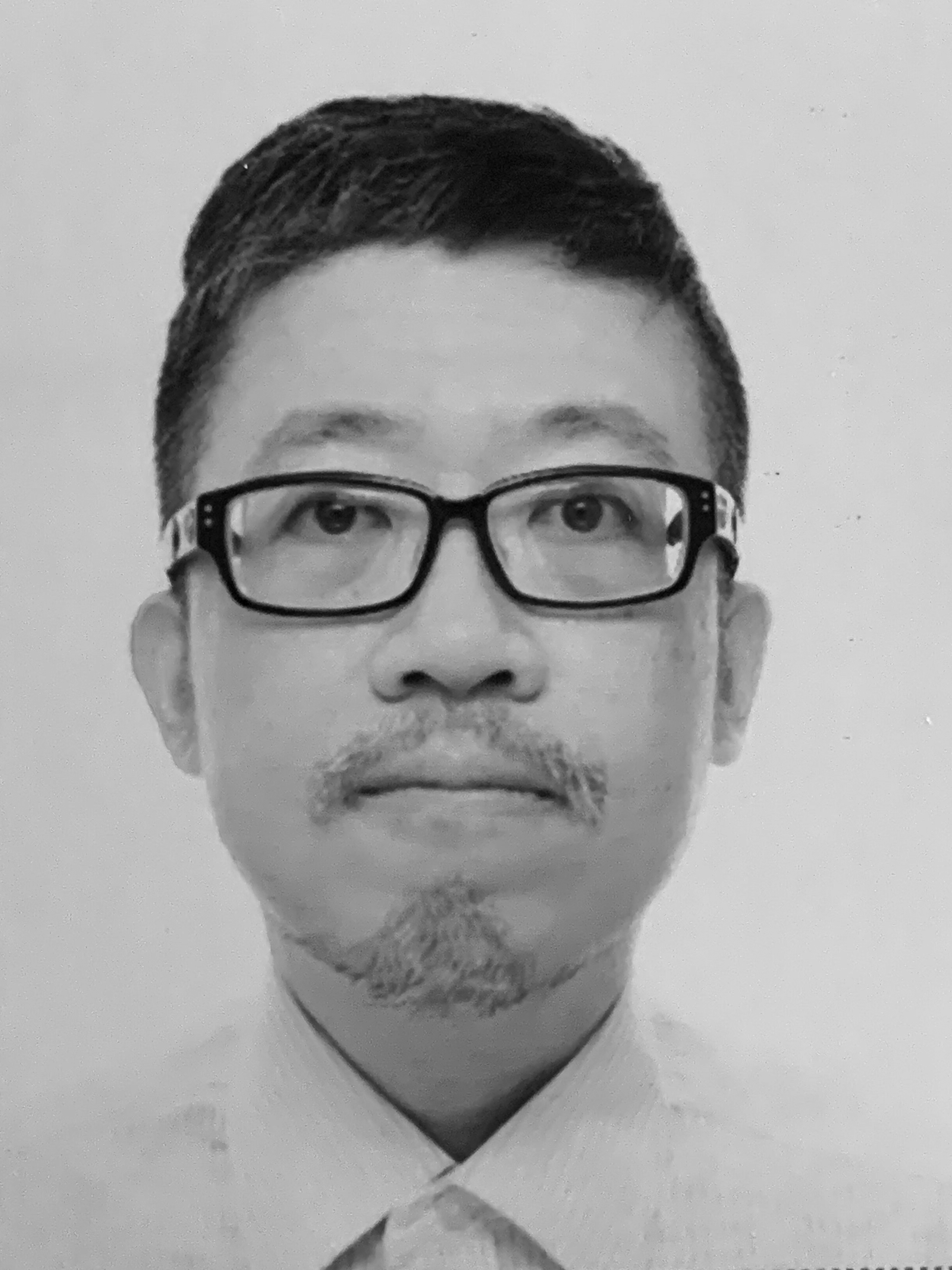}}]
	{Yan-Hsiu Liu}
	received his M.S. degree in Chemistry from National Tsing Hua University (NTHU), Taiwan, in 2002. In 2004, he joined United Microelectronics Corporation (UMC) as a process integration engineer in Hsinchu, Taiwan. He is currently working as a deputy department manager on the development of smart manufacturing and responsible for industry-academia cooperation/collaboration. His research interests include the areas of intelligent manufacturing systems, adaptive parameter estimation, and neural networks.
\end{IEEEbiography}

	
%以下幾行先註解掉,
%要產生listoffigure和listoftable時才會需要用到
\iffalse	
\onecolumn

\listoffigures
\newpage

\listoftables
\newpage
\fi

\end{document}